\documentclass[10pt,twocolumn,letterpaper]{article}

\usepackage{cvpr}              %

\usepackage[dvipsnames]{xcolor}

\usepackage{microtype}
\usepackage{booktabs}

\definecolor{cvprblue}{rgb}{0.21,0.49,0.74}
\usepackage[pagebackref,breaklinks,colorlinks,allcolors=cvprblue]{hyperref}

\def\paperID{4119} %
\def\confName{CVPR}
\def\confYear{2025}

\title{MegaSaM: Accurate, Fast, and Robust Structure and Motion from \\ Casual Dynamic Videos}

\author{
Zhengqi Li$^1$\qquad
Richard Tucker$^1$\qquad
Forrester Cole$^1$\qquad
Qianqian Wang$^{1,2}$\qquad
Linyi Jin$^{1,3}$ \\
Vickie Ye$^2$\qquad
Angjoo Kanazawa$^2$\qquad
Aleksander Holynski$^{1,2}$\qquad
Noah Snavely$^1$
\\[0.5em]
$^1$Google DeepMind \ \ \
$^2$UC Berkeley \ \ \ 
$^3$University of Michigan \ \ \ 
}

\usepackage{mathalfa}
\usepackage{graphicx}
\usepackage{graphbox}
\usepackage{array}
\usepackage{enumitem}
\usepackage{animate}
\usepackage{xspace} 

\definecolor{rowblue}{RGB}{220,230,240}
\definecolor{myorchid}{RGB}{150,10,30}
\definecolor{myblue}{RGB}{10,30,250}
\definecolor{mygreen}{RGB}{10,190,10}
\definecolor{myred}{RGB}{190,20,20}
\definecolor{mypurple}{RGB}{255,100,255}

\newcommand{\newtext}[1]{#1}

\AtBeginDocument{%
 \abovedisplayskip=6pt plus 3pt minus 5pt
 \abovedisplayshortskip=0pt plus 3pt
 \belowdisplayskip=6pt plus 3pt minus 5pt
 \belowdisplayshortskip=5pt plus 3pt minus 4pt
}

\makeatletter
\renewcommand{\paragraph}{%
  \@startsection{paragraph}{4}%
  {\z@}{0.4ex \@plus 1ex \@minus .1ex}{-1em}%
  {\normalfont\normalsize\bfseries}%
}
\makeatother

\makeatletter
\@namedef{ver@everyshi.sty}{} %
\makeatother

\newlength{\itemwidth} %
\usepackage{tikz} %
\usetikzlibrary{calc} %
\usetikzlibrary{tikzmark} %
\usetikzlibrary{spy} %
\usetikzlibrary{shapes.misc} %
\usepackage{pgfplots} %
\usepackage{pgfplotstable} %
\pgfplotsset{compat=newest} %

\newcommand{\method}{MegaSaM\xspace}

\newcommand{\AbsDepth}{D^{\text{abs}}_i}
\newcommand{\RelDepth}{D^{\text{rel}}_i}
\newcommand{\AlignDepth}{D^{\text{align}}_i}

\newcommand{\Wmono}{w_{d}}
\newcommand{\MotionMask}{\hat{m}_i}
\newcommand{\Depth}{\hat{D}}
\newcommand{\DroidDepth}{\hat{\mathbf{d}}}
\newcommand{\DroidDepthi}{\hat{\mathbf{d}}_i}

\newcommand{\Pose}{\hat{\mathbf{G}}}
\newcommand{\Poseij}{\hat{\mathbf{G}}_{ij}}

\newcommand{\focal}{\hat{f}}
\newcommand{\FLowFieldsHat}{{\hat{\mathbf{u}}}_{ij}}
\newcommand{\ConfidenceTilde}{{\tilde{\mathbf{w}}}_{ij}}

\newcommand{\FLowFields}{{\mathbf{u}}_{ij}}

\newcommand{\ConfidenceHat}{\hat{\mathbf{w}}_{ij}}

\newcommand{\RNN}{F}
\newcommand{\RNNm}{F_{m}}

\newcommand{\StateVec}{\boldsymbol{\xi}}

\newcommand{\FrameGraph}{\mathcal{P}}
\newcommand{\MotionPorb}{\mathbf{m}}

\begin{document}
\maketitle
\begin{abstract}

We present a system that allows for accurate, fast, and robust estimation of camera parameters and depth maps from casual monocular videos of dynamic scenes. Most conventional structure from motion and monocular SLAM techniques assume input videos that feature predominantly static scenes with large amounts of parallax. Such methods tend to produce erroneous estimates in the absence of these conditions. Recent neural network-based approaches attempt to overcome these challenges; however, such methods are either computationally expensive or brittle when run on dynamic videos with uncontrolled camera motion or unknown field of view. We demonstrate the surprising effectiveness of a deep visual SLAM framework: with careful modifications to its training and inference schemes, this system can scale to real-world videos of complex dynamic scenes with unconstrained camera paths, including videos with little camera parallax. Extensive experiments on both synthetic and real videos demonstrate that our system is significantly more accurate and robust at camera pose and depth estimation when compared with prior and concurrent work, with faster or comparable running times. See interactive results on our project page: \href{https://mega-sam.github.io/}{mega-sam.github.io}.

\end{abstract}

\section{Introduction}

Extracting camera parameters and scene geometry from a set of images is a fundamental problem in computer vision, commonly referred to as structure from motion (SfM) or Simultaneous Localization and Mapping (SLAM). 
While decades of research have yielded mature algorithms for stationary scenes with large camera baselines, these methods often falter when applied to casual monocular videos captured in uncontrolled setting~\cite{kopf2021rcvd, zhang2022structure}. Such videos, frequently captured by handheld cameras, typically exhibit limited camera motion parallax (e.g., nearly stationary or rotational cameras) and a broad range of focal lengths, and often include moving objects and scene dynamics. 
Recent efforts to address these challenges have focused on two primary strategies: optimizing camera and scene geometry by fine-tuning mono-depth networks or reconstructing radiance fields~\cite{luo2020consistent, zhang2021consistent, zhang2022structure, liu2023robust}; or combining intermediate estimates derived from monocular video, such as depth, flows, long-term trajectory and motion segmentation, into a global optimization framework~\cite{kopf2021rcvd, chen2024leap, zhang2024monst3r, zhao2022particlesfm, liu2023robust}. However, these approaches are computational expensive or brittle when applied to unconstrained videos that feature long time duration, unconstrained camera paths, or complex scene dynamics.

\begin{figure}[tb]
\centering
  \includegraphics[width=1.\columnwidth]{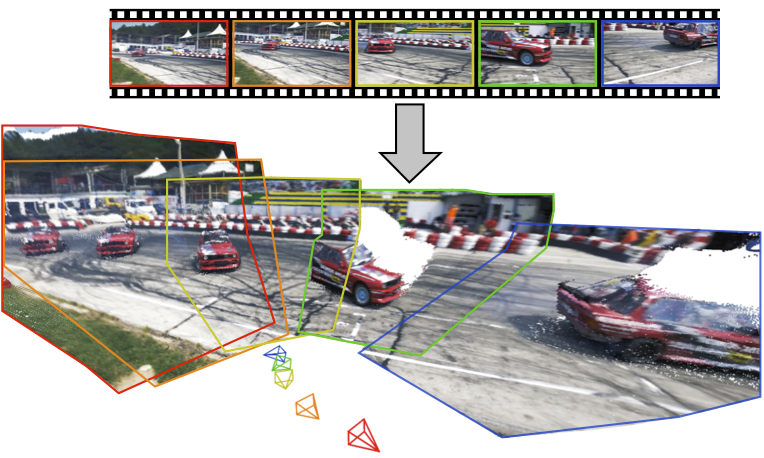} 
  \caption{\textbf{\method} enables accurate, fast and robust estimation of cameras and scene structure from a casually captured monocular video of a dynamic scene. Top: input video frames (every tenth frame shown). Bottom: our estimated camera and 3D point clouds unprojected by predicted video depths without any postprocessing.}
\label{fig:teaser}
\end{figure}

In this work, we present \method, a full
pipeline for \emph{accurate, fast and robust} camera tracking and depth estimation from in-the-wild monocular videos of dynamic scenes. Our approach combines the strengths of several prior works, leading to results with previously unachievable quality, as shown in Fig.~\ref{fig:teaser}.
In particular, we reexamine and extend prior deep visual SLAM framework for camera tracking. A defining feature of deep visual SLAM systems like DROID-SLAM~\cite{teed2021droid} is that they adopt a differentiable bundle adjustment (BA) layer that iteratively updates scene geometry and camera pose variables, and intermediate predictions are learned from large amounts of data through camera and flow supervision. We find that such a learned layer is critical for achieving accurate and efficient camera pose estimation for the more challenging case of dynamic video. Building on this foundation, one of our key innovations for handling dynamic scene is to integrate monocular depth priors and motion probability maps into a differentiable SLAM paradigm.

Further, we analyze the observability of structure and camera parameters in the video and introduce an uncertainty-aware global BA scheme, which improves system robustness when camera parameters are poorly constrained by the input video. 
We also demonstrate how consistent video depths can be accurately and efficiently obtained without the need for test-time network fine-tuning. Extensive evaluation on synthetic and real-world datasets demonstrates that our system significantly outperforms prior and concurrent baselines in both camera and depth estimation accuracy, while achieving competitive or superior runtime performance.

\section{Related Work}

\paragraph{Visual SLAM and SfM.}
SLAM and SfM are used to estimate camera parameters and
3D scene structure from video sequences or unstructured image collections. Conventional approaches tackle this problem by first estimating 2D correspondence between images through feature matching~\cite{snavely2006photo, agarwal2011building, schonberger2016structure, pollefeys2004visual, pollefeys2008detailed, campos2021orb, mur2015orb, davison2007monoslam, klein2007parallel, sweeney2019structure} or photometric alignment~\cite{newcombe2011dtam, engel2014lsd, engel2017direct}. They then optimize 3D point locations and camera parameters by minimizing reprojection or photo-consistency errors through bundle adjustment (BA)~\cite{triggs2000bundle}.  

Recently, deep visual SLAM and SfM systems have emerged that adopt deep neural networks to estimate pairwise or long-term correspondences~\cite{bloesch2018codeslam, tang2018ba, wang2017deepvo, yang2020d3vo, teed2024deep, teed2021droid, wang2024vggsfm, shen2023dytanvo, czarnowski2020deepfactors, he2024detector, kastenfast, holynski2020reducing}, to reconstruct radiance fields~\cite{lin2021barf, Fu_2024_CVPR, park2023camp} or global 3D point clouds~\cite{leroy2024grounding, wang2024dust3r}. 
While these methods demonstrate accurate camera tracking and reconstruction, they typically assume predominantly static scenes and sufficient camera baselines between frames. Therefore, their performance can degrade significantly or fail entirely in the presence of scene dynamics or limited camera parallax.

Several recent works share similar goals with ours in addressing these limitations. Robust-CVD~\cite{kopf2021rcvd} and CasualSAM~\cite{zhang2022structure} jointly estimate camera parameters and dense depth maps from dynamic videos by optimizing a spatially varying spline or fine-tuning monocular depth networks.
Particle-SfM~\cite{zhao2022particlesfm} and LEAP-VO~\cite{chen2024leap} first infer moving object masks based on long-range trajectories, then use this information to downweight the contribution of the features during bundle adjustment. 
Concurrent work, MonST3R~\cite{zhang2024monst3r}, adopts a 3D point cloud representation from DuST3R~\cite{wang2024dust3r} and localizes cameras via an additional alignment optimization. Our approach shares similar ideas, but we show that performance can be significantly improved by coupling the differentiable SLAM system with the intermediate predictions of underlying dynamic scenes.

\paragraph{Monocular depth.}
Recent work on monocular depth prediction has shown strong generalization on in-the-wild single images by training deep neural networks on large amounts of synthetic and real-world data~\cite{li2018megadepth, li2019learning, ranftl2021vision, yin2021learning, ranftl2020towards, saxena2024surprising, depthanything, yang2024depth, yin2023metric3d, piccinelli2024unidepth, ke2024repurposing, godard2017unsupervised}. 
However, these single-image models tend to produce temporally inconsistent 
depth from videos. To overcome this issue, prior techniques propose to fine-tune mono-depth models by performing test-time optimization~\cite{luo2020consistent, zhang2021consistent} or to use transformer or diffusion models to directly predict video depths~\cite{wang2023neural, shao2024learning, hu2024depthcrafter}. Our approach follows the spirit of the first paradigm, but we show that we can achieve better video depth quality without resorting to expensive network fine-tuning for every video.

\paragraph{Dynamic scene reconstruction.}
Several recent works have adopted time-varying radiance field representations to perform dynamic scene reconstruction and novel view synthesis from in-the-wild videos~\cite{park2021nerfies, park2021hypernerf, li2021neural, li2023dynibar, gao2022monocular, wang2024gflow, lei2024mosca, wang2024shape, liu2023robust, wu2024cat4d}.
Our work is orthogonal to most of these techniques since most radiance field reconstruction methods require camera parameters or video depth maps as inputs, and our outputs can be used as inputs to these systems. %

\section{\newtext{MegaSaM}}

Given an unconstrained, continuous video sequence $\mathcal{V} = \{ I_i \in \mathcal{R}^{H \times W} \}^N_{i=1}$ 
our goal is to estimate camera poses $ \Pose_i \in SE(3)$, focal length $\focal$ (if unknown), and dense video depth maps $\mathcal{D} = \{ \Depth_i \}^N_{i=1}$. 
Our approach does not pose any constraints on the camera and object motions present in the input video.
Our camera tracking and video depth estimation modules build upon prior deep visual SLAM (in particular, DROID-SLAM~\cite{teed2021droid}) and casual structure and motion~\cite{zhang2022structure} frameworks, respectively.

In the following section, we first summarize the key components of a deep visual SLAM framework designed for tracking videos of static scenes with sufficient camera motion parallax (Sec.~\ref{sec:formulation}). 
We then introduce key modifications to this framework, at both the training and inference stages, that enable fast, robust and accurate camera tracking for unconstrained dynamic videos (Sec.~\ref{sec:extension}). Finally, we demonstrate how to efficiently estimate consistent video depths given the estimated camera parameters (Sec.~\ref{sec:video-depth}).

\subsection{Deep visual SLAM formulation} \label{sec:formulation}

Deep visual SLAM systems like DROID-SLAM~\cite{teed2021droid} are characterized by a differentiable, learned bundle adjustment (BA) layer that iteratively updates structure and motion parameters. 
In particular, they keep track of two state variables while processing a video: 
a per-frame low-resolution disparity map $\DroidDepthi \in \mathcal{R}^{\frac{H}{8} \times \frac{W}{8}}$, and camera poses
$\Pose_i \in SE(3)$. 
These variables are updated iteratively during both training and inference stages through the differentiable BA layer, which operates over a set of image pairs from a frame-graph $(I_i, I_j) \in \FrameGraph$, built dynamically to connect frames with overlapping field of view.

Given two video frames $I_i$ and $I_j$ from the frame graph as input, DROID-SLAM
learns to predict a 2D correspondence field $\FLowFieldsHat \in \mathcal{R}^{\frac{H}{8} \times \frac{W}{8} \times 2}$ and a confidence $\ConfidenceHat \in \mathcal{R}^{\frac{H}{8} \times \frac{W}{8}}$ through  convolutional gated recurrent units in an iterative manner:
$(\FLowFieldsHat^{k+1}, \ConfidenceHat^{k+1}) = \RNN(I_i, I_j, 
\FLowFieldsHat^{k}, \ConfidenceHat^{k})$, where $k$ denotes the $k^{th}$ iteration. In addition,
the rigid-motion correspondence field can also be derived from the camera ego-motion and the disparity through a multi-view constraint:
\begin{equation}
    {\FLowFields} = \pi \left( \Poseij \circ \pi^{-1} (\mathbf{p}_i, \DroidDepthi, K^{-1}) , K \right),
\end{equation}
where $\mathbf{p}_i$ denotes a grid of pixel coordinates, $\pi$ denotes the perspective projection operator, $\Pose_{ij} = \Pose_{j} \circ \Pose^{-1}_{i}$ is the relative camera pose between $I_i$ and $I_j$, and $K \in \mathcal{R}^{3 \times 3}$ denotes the camera intrinsic matrix.

\medskip
\noindent \textbf{Differentiable bundle adjustment.}
DROID-SLAM assumes known focal length, but focal length is typically not known a priori for in-the-wild videos. 
Therefore, we optimize camera poses, focal length and disparity by iteratively minimizing a weighted reprojection cost between the current flows predicted by the network and the rigid-motion ones derived from camera parameters and disparity~\cite{hagemann2023deep}:
\begin{equation}
    \mathcal{C}(\Pose, \DroidDepth, \focal) = \sum_{(i, j) \in \FrameGraph} || \FLowFieldsHat - \FLowFields ||_{\Sigma_{ij}}^2 \label{eq:ba_1}
\end{equation}
where weights $\Sigma_{ij}=\text{diag}(\ConfidenceHat)^{-1}$.
To enable differentiable end-to-end training, we perform optimization of Eq.~\ref{eq:ba_1} with the Levenberg–Marquardt algorithm:
\begin{equation}
    \left( \mathbf{J}^T \mathbf{W} \mathbf{J} + \lambda \text{diag}({\mathbf{J}^T \mathbf{W} \mathbf{J}}) \right) \Delta \StateVec =  \mathbf{J}^T \mathbf{W} \mathbf{r} \label{eq:lm}
\end{equation}
where $\Delta \StateVec = (\Delta \mathbf{G}, \Delta \mathbf{d}, \Delta f)^T$ is the parameter updates of the state variables, $\mathbf{J}$ is the Jacobian of reprojection residuals w.r.t. the parameters, and $\mathbf{W}$ is a diagonal matrix containing $\ConfidenceHat$ from each frame pair.
$\lambda$ is a damping factor predicted by the network during each BA iteration.
We can separate camera parameters (including poses and focal length) and disparity variables by dividing the approximated Hessian on the LHS of Eq.~\ref{eq:lm} into following block matrix form:
\begin{align}
\begin{bmatrix}
\mathbf{H}_{\mathbf{G}, f} & \mathbf{E}_{\mathbf{G}, f} \\
\mathbf{E}_{\mathbf{G}, f}^T & \mathbf{H}_{\mathbf{d}} 
\end{bmatrix} 
\begin{bmatrix}
\Delta \StateVec_{\mathbf{G}, f}  \\
\Delta \mathbf{d} \\
\end{bmatrix} = 
\begin{bmatrix}
\tilde{r}_{\mathbf{G}, f} \\
\tilde{r}_{\mathbf{d}} \\
\end{bmatrix} \label{eq:lm2}
\end{align}
Since only a single disparity variable is included in each pairwise reprojection term from Eq.~\ref{eq:ba_1}, $\mathbf{H}_{\mathbf{d}}$ in Eq.~\ref{eq:lm2} is a diagonal matrix, and hence we can efficiently compute parameter updates using the Schur complement trick~\cite{triggs2000bundle}, which leads to fully differentiable BA update:
{\small
\begin{align}
    \Delta \boldsymbol{\xi}_{\mathbf{G}, f}  & = \left[ \mathbf{H}_{\mathbf{G}, f} - \mathbf{E}_{\mathbf{G}, f} \mathbf{H}_{\mathbf{d}}^{-1} {\mathbf{E}_{\mathbf{G} , f}}^{T} \right]^{-1} (\tilde{r}_{\mathbf{G}, f} - \mathbf{E}_{\mathbf{G}, f} \mathbf{H}_{\mathbf{d}}^{-1} \tilde{r}_{\mathbf{d}}) \label{eq:schur1} \\
    \Delta \mathbf{z} &= \mathbf{H}_{\mathbf{d}}^{-1} (\tilde{r}_{\mathbf{d}} - \mathbf{E}_{\mathbf{G}, f}^T \Delta \boldsymbol{\xi}_{\mathbf{G}, f}) \label{eq:schur2}
\end{align}
}

\noindent \textbf{Training.}
The flow and uncertainty predictions are trained end-to-end from a collection of synthetic video sequences of static scenes:
\begin{align}
    \mathcal{L}_{\text{static}} = \mathcal{L}_{\text{cam}} + w_{\text{flow}} \mathcal{L}_{\text{flow}} \label{eq:static-loss}
\end{align}
where $\mathcal{L}_{\text{cam}}$ and $\mathcal{L}_{\text{flow}}$ are losses comparing estimated camera parameters and ego-motion induced flows from the BA layer with corresponding ground truths.

\begin{figure}[tb]
\centering
  \includegraphics[width=1.0\columnwidth]{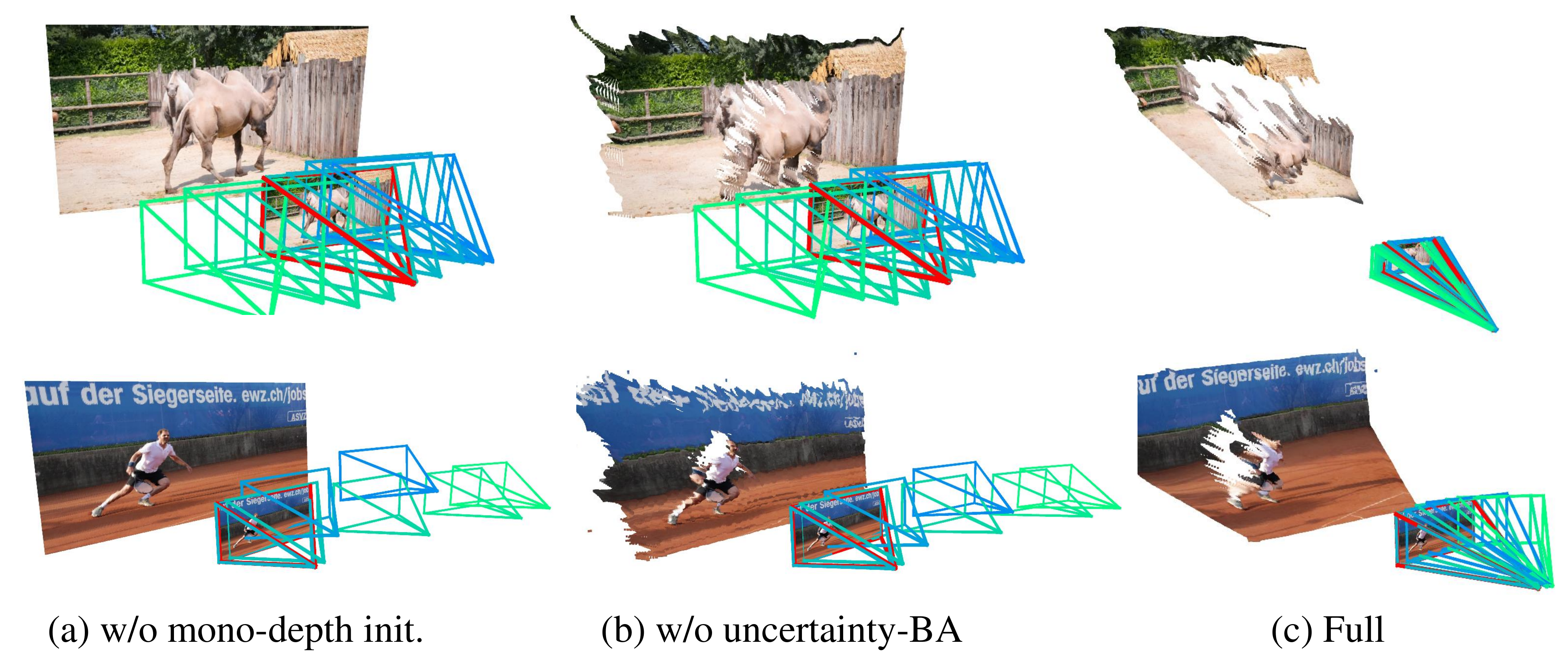} 
  \vspace{-1em}
  \caption{\textbf{Ablation on our design choices.} From left to right, we visualize cameras and reconstruction from our system (a) without mono-depth initialization, (b) without uncertainty-aware BA, (c) with full configuration. For these difficult near-rotational sequences, our full method produces much better camera and scene geometry.} %
\label{fig:ablation}
\end{figure}

\subsection{Scaling to in-the-wild dynamic videos} \label{sec:extension} 

Deep Visual SLAM works reasonably well for videos that feature static scenes and have sufficient camera translation, but its performance degrades when operating on videos of dynamic content, or videos with limited parallax,
as shown in the first column of Fig.~\ref{fig:ablation}. To overcome these issues, we propose key modifications to the original training and inference pipeline.
First, our model predicts object movement maps that are learned together with flow and uncertainty in order to downweight dynamic elements within the differentiable BA layer. Second, we propose to integrate priors from mono-depth estimates into the both training and inference pipeline, and perform uncertainty-aware global BA, both of which aid disambiguation of object and camera motion in challenging casual dynamic videos. Our system is trained on synthetic data alone, but we demonstrate its strong generalization to real-world videos.

\subsubsection{Training}

\noindent \textbf{Learning motion probability.}
Recall from Sec.~\ref{sec:formulation} that for every selected image pair $(I_i, I_j) \in \FrameGraph$, our model predicts a 2D flow $\FLowFieldsHat$ and associated confidence $\ConfidenceHat$ at each BA iteration, and that these predictions are supervised from synthetic sequences of static scenes. 
To extend the model to handle dynamic scenes, 
we can directly train the model predictions on videos of dynamic scenes with corresponding ground truth supervision, hoping that the pairwise uncertainty will subsume the object motion information automatically during training. 
However, we find that this simple training strategy tends to produce suboptimal results due to unstable training behaviour from differentiable BA layers.

Instead, we propose to use an additional network $\RNNm$ to iteratively predict an \emph{object movement probability map} $\MotionPorb_i \in \mathcal{R}^{\frac{H}{8} \times \frac{W}{8}} = \RNNm \left( \{ I_i \} \cup \mathcal{N}(i) \right)$ conditioned on $I_i$ and a set of its neighboring keyframes $\mathcal{N}(i) = \{ I_j | (i, j) \in \FrameGraph \}$. This movement map is specifically supervised to predict pixels that correspond to dynamic content based on multi-frame information. During each BA iteration, we combine pairwise flow confidence $\ConfidenceHat$ with object movement map $\MotionPorb_i$ to form the final weights in Eq.~\ref{eq:ba_1}: $\ConfidenceTilde = \ConfidenceHat \MotionPorb_i$.

Moreover, we design a two-stage training scheme that trains the models on a mixture of static and dynamic videos to effectively learn 2D flows along with the movement probability maps.
In the first \emph{ego-motion pretraining} stage, we train the original deep SLAM model $\RNN$ by supervising the predicted flows and confidence maps (using the losses in Eq.~\ref{eq:static-loss}) with synthetic data of static scenes, \textit{i.e.}, without any dynamic video data. This stage helps model effectively learn pairwise flows and corresponding confidence induced only by ego-motion.
In the second  \emph{dynamic fine-tuning} stage, we freeze the parameters of $\RNN$ and finetune $\RNNm$ on synthetic dynamic videos, conditioning $\RNNm$ on the features from our pretrained $\RNN$ during each iteration to predict movement probability map $\MotionPorb_i$, supervising through both camera and cross-entropy losses:
\begin{equation}
    \mathcal{L}_{\text{dynamic}} = \mathcal{L}_{\text{cam}} + w_{\text{motion}} \mathcal{L}_{\text{CE}}
\end{equation}

\begin{figure}[tb]
    \centering
    \setlength{\tabcolsep}{0.01cm}
    \begin{tabular}{cc}
        \includegraphics[width=0.45\columnwidth]{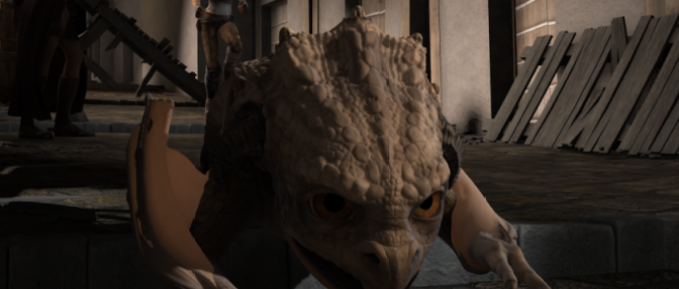} &
        \includegraphics[width=0.45\columnwidth]{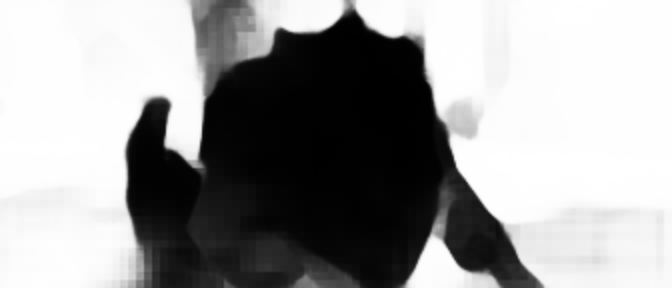} \\
        \includegraphics[width=0.45\columnwidth]{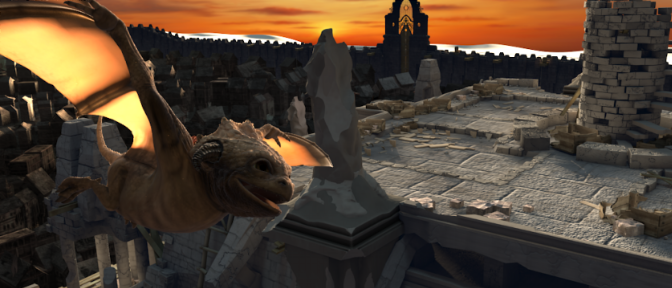} &
        \includegraphics[width=0.45\columnwidth]{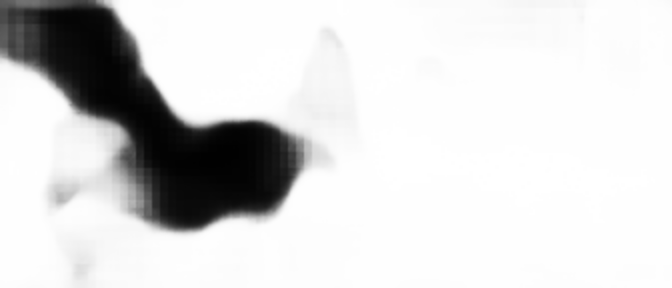} \\
        \vphantom{I} Image &
        \vphantom{I} Motion probability
    \end{tabular}
    \vspace{-0.3cm}
    \caption{\textbf{Learned movement map.} Left: input video frame, right: corresponding learned motion probability map.} \label{fig:movement_map}
\end{figure}

\noindent This stage decorrelates learning scene dynamics from learning 2D correspondences, and thus leads to more stable and effective training behavior for the differentiable BA framework.
We found this training scheme to be critical for producing accurate camera estimation results for dynamic videos, as shown in our ablation study. We visualize learned motion probability maps $\MotionPorb_i$ in Fig.~\ref{fig:movement_map}.

\smallskip
\noindent \textbf{Disparity and camera initialization.}
DROID-SLAM initializes disparity $\DroidDepth$ by simply setting it to a constant value of $1$. However, we find that this initialization fails to perform accurate camera tracking on videos with limited camera baselines and complex scene dynamics. Inspired by recent work~\cite{li2021neural, wang2024shape, li2023dynibar, liu2023robust}, during both the training and inference stages, we perform data-driven initialization by integrating a mono-depth prior.
During training, we initialize $\DroidDepth$ with disparity from DepthAnything~\cite{depthanything} with the estimated global scale and shift borrowed from the ground truth depth of each training sequence. For each training sequence, we initialize first two camera poses to the ground-truth to remove the gauge ambiguity, and initialize the camera focal length by randomly perturbing the ground truth value by $25\%$.

\subsubsection{Inference}

Our inference pipeline consists of two components: (i) a frontend module registers cameras for keyframes by performing frame selection followed by a sliding window BA. (ii) a backend module refines the estimates by performing global BA over all video frames. In this subsection, we describe the modifications we made at inference.

\medskip
\noindent \textbf{Initialization and frontend tracking.}
Similar to training, we incorporate both mono-depth and focal length predictions into the inference pipeline. In particular, we initialize per-frame disparity maps $\DroidDepth_i$ with metric aligned monocular disparity $\AlignDepth = \hat{\alpha} \RelDepth + \hat{\beta}$, where $\RelDepth$ is per-frame affine-invariant disparity from~\cite{depthanything}, and per-video global scale and shift parameters $(\hat{\alpha}, \hat{\beta})$ are estimated through median aligning $\RelDepth$ with  
an additional metric depth estimates $\AbsDepth$ from UniDepth~\cite{piccinelli2024unidepth}:
$\hat{\alpha}_i=\frac{\AbsDepth - \text{median}_i(\AbsDepth)}{\RelDepth - \text{median}(\RelDepth)}$; $\hat{\beta}= \text{median} \left( \AbsDepth - \hat{\alpha} \RelDepth \right)$.\footnote{We combine two mono-depth models since they have complementary advantages: DepthAnything provides more accurate and consistent depths, whereas UniDepth provides scene scale and shift estimates.}
The UniDepth model also predicts the focal length of each frame; we use the median estimate across the video frames to obtain an initial focal length estimate $\hat{f}$, which is fixed within the frontend stage. 

To initialize the SLAM system, we accumulate keyframes with sufficient pairwise motion until we have a set of $N_{\text{init}}=8$ active images. We initialize camera poses for these keyframes by performing camera motion--only bundle adjustment while fixing disparity variables $\DroidDepth_i$ to aligned monocular disparity $\AlignDepth$. After initialization, we incrementally add new keyframes, remove old keyframes, and perform local BA in a sliding window manner, where each keyframe disparity is also initialized to aligned mono-disparity. 
In this stage, the BA cost function consists of a reprojection error and a mono-depth regularization term:
{
\small
\begin{equation}
    \mathcal{C}  = \sum_{(i, j) \in \mathcal{P}} || \FLowFieldsHat - \FLowFields ||_{\Sigma_{ij}}^2 + \Wmono \sum_{i}|| \DroidDepth_i  - \AlignDepth ||^2. \label{eq:ba_mono}
\end{equation}
}

\begin{figure}[tb]
\centering
  \includegraphics[width=1.0\columnwidth]{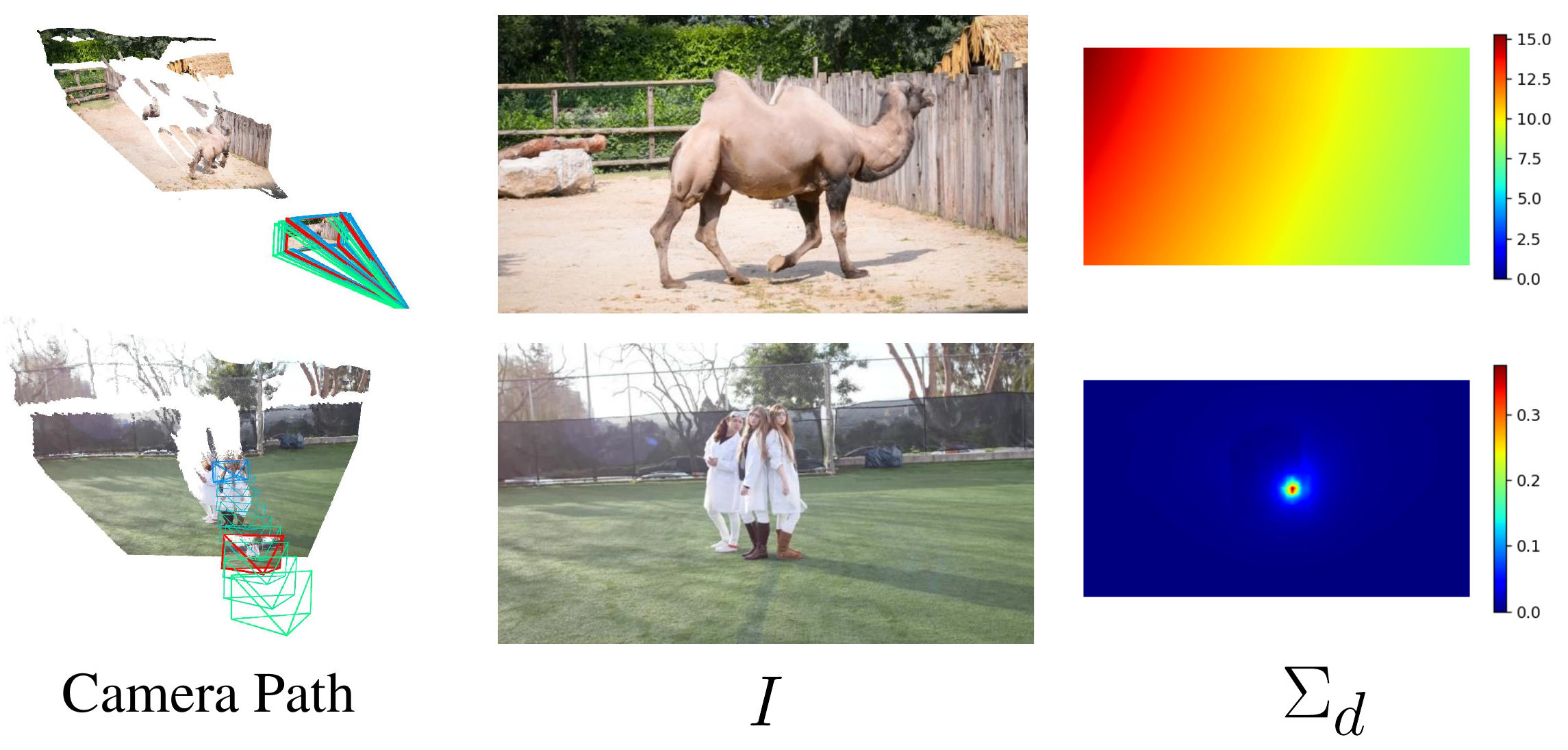} 
    \caption{\textbf{Visualization of epistemic uncertainty.} From left to right, we visualize camera paths, reference image and corresponding epistemic uncertainty of disparity. 
    The geometry is not observable from the top example with little camera parallax, as indicated by the larger uncertainty.
    The peak on the bottom uncertainty map corresponds to the epipole for forward moving motion.} \label{fig:uncertainty}
\end{figure}

\medskip
\noindent \textbf{Uncertainty-aware global BA.}
The backend module first performs global BA over all keyframes. The module then performs a pose-graph optimization to register the poses of non-keyframes.
Finally, the backend module refines the entire camera trajectory via a global BA over all video frames.

This design raises a question: should we (or when should we) add mono-depth regularization from Eq.~\ref{eq:ba_mono} into the global bundle adjustment?
One on hand, if there is sufficient camera baseline in the input video, we observe that mono-depth regularization isn't needed, since the problem is already well-constrained 
and in fact the error from mono-depth can diminish camera tracking accuracy. 
On the other hand, if a video is captured from a rotational camera with little camera baseline, then performing a reprojection-only BA without additional constraints can lead to degenerate solutions, as shown in the second column of Fig.~\ref{fig:ablation}.

To see why, 
we explore the approximate Hessian matrix from the linear system in Eq.~\ref{eq:lm2}. 
As shown by Goli~\etal~\cite{goli2024bayes}, 
given the posterior, $p(\theta | \mathcal{I})$, we can use Laplace approximations to estimate the covariance $\Sigma$ of the variables through the inverse Hessian: $\Sigma_{\theta} = - \mathbf{H}(\theta^{*})^{-1}$, where $\theta^{*}$ is the MAP estimate of the parameters and $\Sigma_{\theta}$ denotes the epistemic uncertainty of estimated variables~\cite{kendall2017uncertainties}. Since inverting the full Hessian is computationally expensive when the number of input frames is large, we follow Ritter~\etal~\cite{ritter2018scalable} and approximate $\Sigma_{\theta}$ through the diagonal of the Hessian:
\begin{equation}
    \Sigma_{\theta} \approx \text{diag}\left( -\mathbf{H}(\theta^{*}) \right)^{-1}
\end{equation}

Intuitively, when we consider the reprojection error in Eq.~\ref{eq:ba_1}, the Jacobian of estimated variables $\mathbf{J}_{\theta}$ indicates how much the reprojection error would change if we perturb the variables. 
Therefore, the uncertainty $\Sigma_{\theta}$ is large when perturbing the parameters has little impact on the reprojection error.
Specifically, let us consider the disparity variables, and consider an extreme case where the input video is captured by a static camera. In this case, the pairwise reprojection error would be unchanged as a function of disparity, implying a large uncertainty in the estimated disparity; that is, the disparity is unobservable from the video alone. We visualize the spatial uncertainty of the estimated normalized disparity $\Sigma_{d}$
in Fig~\ref{fig:uncertainty}: the first row shows a video featuring rotational dominant motion, whereas the second row shows videos captured by a forward-moving camera. From the color bars in the third column, we see the range of disparity uncertainty $\Sigma_{d}$ is much higher in the first example.

Such uncertainty quantification gives us a measure for observability of 
camera and disparity parameters, allowing us to decide where we should add mono-depth regularization (and additionally when we should turn off camera focal length optimization). 
In practice, we find that simply checking the median uncertainty of normalized disparity and uncertainty of normalized focal length works well for all the videos we tested.
In particular, after completing frontend tracking, we retrieve the diagonal entry of disparity Hessian formed from all the keyframes and compute its median $\text{med} \left(\text{diag} (\mathbf{H}_{\mathbf{d}}) \right)$, as well as the Hessian entry of the shared focal length ${H}_{f}$.
We then set the mono-depth regularization weight based on the median disparity Hessian $\Wmono = \gamma_d \exp \left( - \beta_d \text{med} \left(\text{diag} (\mathbf{H}_{\mathbf{d}}) \right) \right)$.
In other words, we enable mono-depth regularization if the camera poses are unobservable from the input video alone due to limited camera motion parallax. 
In addition, we disable focal length optimization if ${H}_{f} < \tau_f$, since this condition indicates that focal length is likely unobservable from the input.

\subsection{Consistent depth optimization} \label{sec:video-depth} 
Optionally, one can obtain more accurate and consistent video depth at higher resolution than the estimated low-res disparity variables given the estimated camera parameters.

In particular, we follow 
CasualSAM~\cite{zhang2022structure} and perform an additional first-order optimization on video depths along with per-frame aleatoric uncertainty maps. Our objective consist of three cost functions:
\begin{equation}
    \mathcal{C}_{\text{cvd}} = w_{\text{flow}} \mathcal{C}_{\text{flow}} +  w_{\text{temp}} \mathcal{C}_{\text{temp}} + w_{\text{prior}} \mathcal{C}_{\text{prior}}
\end{equation}
where $\mathcal{C}_{\text{flow}}$ denotes pairwise 2D flow reprojection loss, $\mathcal{C}_{\text{temp}}$ is temporal depth consistency loss, and  $\mathcal{C}_{\text{prior}}$ is scale invariant mono-depth prior loss. We derive 2D optical flow at the original frame resolution from an off-the-shelf module~\cite{teed2020raft}. %

Note that our design has a few differences compared to CasualSAM: (i) instead of performing a time-consuming mono-depth network fine-tuning, we construct and optimize a sequence of variables for disparity and uncertainty over the input video; 
(ii) we fix camera parameters instead of jointly optimizing cameras and depths during optimization; 
(iii) we adopt surface normal consistency and multi-scale depth gradient matching losses~\cite{li2018megadepth, sayed2022simplerecon} to replace the depth prior loss used in CasualSAM~\cite{zhang2022structure}. 
We find that these modifications lead to much faster optimization times as well as more accurate video depth estimates. We refer readers see the supplementary material for more details of our losses and optimization scheme.

\section{Experiments}  \label{sec:experiment}

\setlength{\tabcolsep}{2.2pt}
\begin{table}[t]
\begin{center}
\small
\begin{tabular}{l @{\hskip 0.5em} ccc @{\hskip 0.5em} ccc @{\hskip 0.5em} c}
\toprule
& \multicolumn{3}{c@{\hskip 1em}}{Calibrated} &
\multicolumn{3}{c@{\hskip 1em}}{Uncalibrated} &
\\
Method & 
ATE & RTE & RRE & ATE & RTE & RRE & Time
\\ 
\midrule
CasualSAM~\cite{zhang2022structure} & 0.036 & 0.013 & 0.20 & 0.067 & 0.019 & 0.47 & 1.6m \\
LEAP-VO~\cite{chen2024leap} & 0.041 & 0.023 & 0.17 & - & - & - &  1.3s \\
ACE-Zero~\cite{brachmann2024scene} & 0.053	& 0.028 & 0.30 & 0.065 &	0.028 & 1.92 & 10s \\
Particle-SfM~\cite{zhao2022particlesfm} & 0.062 &	0.032 & 1.26 & 0.057 &	0.038 &	1.64 & 21s \\
RoDynRF~\cite{liu2023robust} & 0.110 & 0.049 &	1.68 & 0.109 &	0.051 &	1.32 & 15m \\
MonST3R~\cite{zhang2024monst3r} & - & - & - & 0.078 & 0.038 & 0.49 & \textbf{1.0s} \\
\bottomrule
Ours & \textbf{0.018} & \textbf{0.008} & \textbf{0.04} & \textbf{0.023} & \textbf{0.008} & \textbf{0.06} & \textbf{1.0s} \\
\end{tabular}
\caption{{Quantitative comparisons of camera estimation on the Sintel dataset.}} \label{table:sintel} 
\end{center} 
\vspace{-1.5em}
\end{table}

\setlength{\tabcolsep}{2.2pt}
\begin{table}[t]
\begin{center}
\small
\begin{tabular}{l ccc @{\hskip 1em} ccc @{\hskip 0.5em} c}
\toprule
& \multicolumn{3}{c@{\hskip 1em}}{Calibrated} &
\multicolumn{3}{c@{\hskip 1em}}{Uncalibrated} &
\\
Method & 
ATE & RTE & RRE & ATE & RTE & RRE & Time
\\ 
\midrule
CasualSAM~\cite{zhang2022structure} & 0.185	& 0.022 & 0.23 & 0.209 & 0.027 & 0.28 & 2.8m  \\
LEAP-VO~\cite{chen2024leap} & 0.167 & 0.011 & 0.09 & - & - & - &  \textbf{0.8s} \\
ACE-Zero~\cite{brachmann2024scene} & 0.062 & 0.012 & 0.11 & 0.056 & 0.012 & 0.12 & 1.6s \\
Particle-SfM~\cite{zhao2022particlesfm} & 0.081 & 	0.014 & 0.20 & 0.087	& 0.015 & 0.29 & 35s \\
RoDynRF~\cite{liu2023robust} & 0.548 &	0.074 &	0.70 & 0.562	& 0.087 & 0.90 & 6.6m\\
MonST3R~\cite{zhang2024monst3r} & - & - & - & 0.690 & 0.078 & 0.54 & 1.0s  \\
\bottomrule
Ours & \textbf{0.020} & \textbf{0.005} & \textbf{0.05} & \textbf{0.020} & \textbf{0.005} & \textbf{0.06} & \textbf{0.8s}
\end{tabular}
\caption{{Quantitative comparisons of camera estimation on the DyCheck dataset.}} \label{table:dycheck}
\vspace{-1.5em}
\end{center} 
\end{table}

\noindent \textbf{Implementation details.} 
In our two-stage training scheme, we first pretrain our model on synthetic data of static scenes,
which include 163 scenes from TartanAir~\cite{tartanair2020iros} and 5K videos from static Kubric~\cite{greff2021kubric}. In the second stage, we finetune motion module $\RNNm$ on 11K dynamic videos from Kubric~\cite{greff2021kubric}. 
Each training example consists of a 7-frame video sequence. 
We set $w_{\text{flow}} = 0.02, w_{\text{motion}} = 0.1$ during training.
Training the camera tracking module using the Adam optimizer~\cite{Kingma2014AdamAM} takes around 4 days with 8 Nvidia 80G A100s. 
Within the initialization and frontend phase, we set mono-depth regularization weight $w_d = 0.05$. In the backend phase, we set $\gamma_d = 1 \times 10^{-4}, \beta_d = 0.05, \tau_f = 50$, 
In term of consistent video depth optimization, we set $w_{\text{flow}} = w_{\text{prior}} = 1.0, w_{\text{temp}} = 0.2$. 
The average running time of our optimization is 1.3 FPS for video depths at resolution $336 \times 144$ on Sintel, but we visualize and evaluate them at resolution of $672 \times 288$.
We refer readers to supplemental material for more details of network architectures, and other training/inference settings.

\noindent \textbf{Baseline.}
We compare \method to recent camera pose estimation methods on both calibrated (known focal length) and uncalibrated (unknown focal length) videos. 
ACE-Zero~\cite{brachmann2024scene} is a state-of-the art camera localization method based on scene coordinate regression, designed for static scenes. 
CasualSAM~\cite{zhang2022structure} and RoDynRF~\cite{liu2023robust} jointly estimate camera parameters and dense scene geometry through optimizing mono-depth networks or instant-NGP~\cite{mueller2022instant}.
Particle-SfM~\cite{zhao2022particlesfm} and LEAP-VO~\cite{chen2024leap} estimate cameras from dynamic videos by predicting motion segmentation from long-term trajectories, then using them to mask out moving objects within a standard visual odometry or SfM pipeline.  
The concurrent work MonST3R~\cite{zhang2024monst3r} extends Dust3R~\cite{wang2024dust3r} to handle dynamic scenes, estimating camera parameters from global 3D point clouds predicted from pairs of input frames. 
To evaluate depth accuracy, we compare our outputs to those from CasualSAM, MonST3R, and VideoCrafter~\cite{hu2024depthcrafter}. We also include raw mono-depth from DepthAnything-V2~\cite{ yang2024depth} for the sake of completeness. 
We run all the above baselines using their respective open-source implementations on the same machine with single Nvidia A100 GPU.

\subsection{Benchmarks and metrics}

\noindent \textbf{MPI Sintel.} 
The MPI Sintel~\cite{Butler:ECCV:2012} dataset includes animated video sequences consisting of complex object motions and camera paths. Following CasualSAM~\cite{zhang2022structure}, we evaluate all methods on the 18 sequences from the dataset, each of which consists of 20-50 images.

\medskip
\noindent \textbf{DyCheck.} 
The DyCheck dataset~\cite{gao2022monocular} was initially designed for evaluating the task of novel view synthesis, and includes real-world videos of dynamic scenes captured from hand-held cameras. 
Each video includes 180-500 frames. 
We use the refined camera parameters and sensor depths provided by Shape of Motion~\cite{wang2024shape} as ground truth.

\setlength{\tabcolsep}{2.2pt}
\begin{table}[t]
\begin{center}
\small
\begin{tabular}{l ccc @{\hskip 1em} ccc @{\hskip 0.5em} c}
\toprule
& \multicolumn{3}{c@{\hskip 1em}}{Calibrated} &
\multicolumn{3}{c@{\hskip 1em}}{Uncalibrated} &
\\
Method & 
ATE & RTE & RRE & ATE & RTE & RRE & Time
\\ 
\midrule
CasualSAM~\cite{zhang2022structure} & 0.031 &	0.005	& 0.31 & 0.035 &	0.005 & 0.30 & 1.1m \\
LEAP-VO~\cite{chen2024leap} & 0.016 & 0.004 &	0.04 & - & - & - &  \textbf{0.6s} \\
ACE-Zero~\cite{brachmann2024scene} & 0.091 & 0.008 & 0.08 & 0.091	& 0.008 & 0.09 & 4.0s \\
Particle-SfM~\cite{zhao2022particlesfm} & 0.051 &	0.007 &	0.10 & 0.054 & 0.007 &	0.14  & 49s \\
RoDynRF~\cite{liu2023robust} & 0.116 &	0.021 &	0.34 & 0.112	& 0.031 & 0.39 & 7.6m \\
MonST3R~\cite{zhang2024monst3r} & - & - & - & 0.073 & 0.014 & 0.18 & 1.7s \\
\bottomrule
Ours & \textbf{0.004} & \textbf{0.001} & \textbf{0.02} & \textbf{0.004} & \textbf{0.001} & \textbf{0.02} & 0.7s
\end{tabular}
\caption{{Quantitative comparisons of camera estimation on a dataset of In-the-Wild footage.} } \label{table:dynibar} 
\end{center} 
\vspace{-1.5em}
\end{table}

\begin{table}[t]
\begin{center}
\small
\begin{tabular}{l ccc @{\hskip 0.5em} ccc}
\toprule
& \multicolumn{3}{c@{\hskip 0.5em}}{Sintel~\cite{Butler:ECCV:2012}} &
\multicolumn{3}{c@{\hskip 0.5em}}{Dycheck~\cite{gao2022monocular}}
\\
Method & 
abs-rel & log-rmse & $\delta_{1.25}$  & abs-rel & log-rmse & $\delta_{1.25}$
\\ 
\midrule
DA-v2~\cite{yang2024depth} & 0.37 & 0.55 & 58.6 & 0.20 & 0.27 & 84.7\\
DepthCrafter~\cite{hu2024depthcrafter} & 0.27 & 0.50 & 68.2 & 0.22 & 0.29 & 83.7 \\
CasualSAM~\cite{zhang2022structure} & 0.31 & 0.49 &	64.2 & 0.21 & 0.30 &	78.4  \\
MonST3R~\cite{zhang2024monst3r} & 0.31 & 0.43 & 62.5 & 0.26 & 0.35 & 66.5 \\
\bottomrule
Ours & \textbf{0.21} & \textbf{0.39} &	\textbf{73.1} & \textbf{0.11} & \textbf{0.20} &	\textbf{94.1}
\end{tabular}
\vspace{-0.5em}
\caption{\textbf{Quantitative comparisons of video depths.} Lower is better for abs-rel and log-rmse, and higher is better for $\delta_{1.25}$.} \label{table:sintel_depth} 
\vspace{-1.5em}
\end{center} 
\end{table}

\medskip
\noindent \textbf{In-the-wild.} 
We further evaluate on in-the-wild dynamic videos. 
Specifically, we include comparisons on 12 in-the-wild videos used by DynIBaR~\cite{li2023dynibar}. These videos feature long time duration (100-600 frames), uncontrolled camera paths, and complex scene motions. 
We construct ground truth movement masks via instance segmentation~\cite{he2017mask}, where the instance IDs are manually specified, and use them to mask out moving objects before running COLMAP~\cite{schonberger2016structure} to obtain reliable camera parameters.

\begin{figure}[tb]
    \centering
    \setlength{\tabcolsep}{0.00cm}
    \setlength{\itemwidth}{2.8cm}
    \renewcommand{\arraystretch}{0.1}
    \hspace*{-\tabcolsep}\begin{tabular}{ccc}
            \includegraphics[width=\itemwidth]{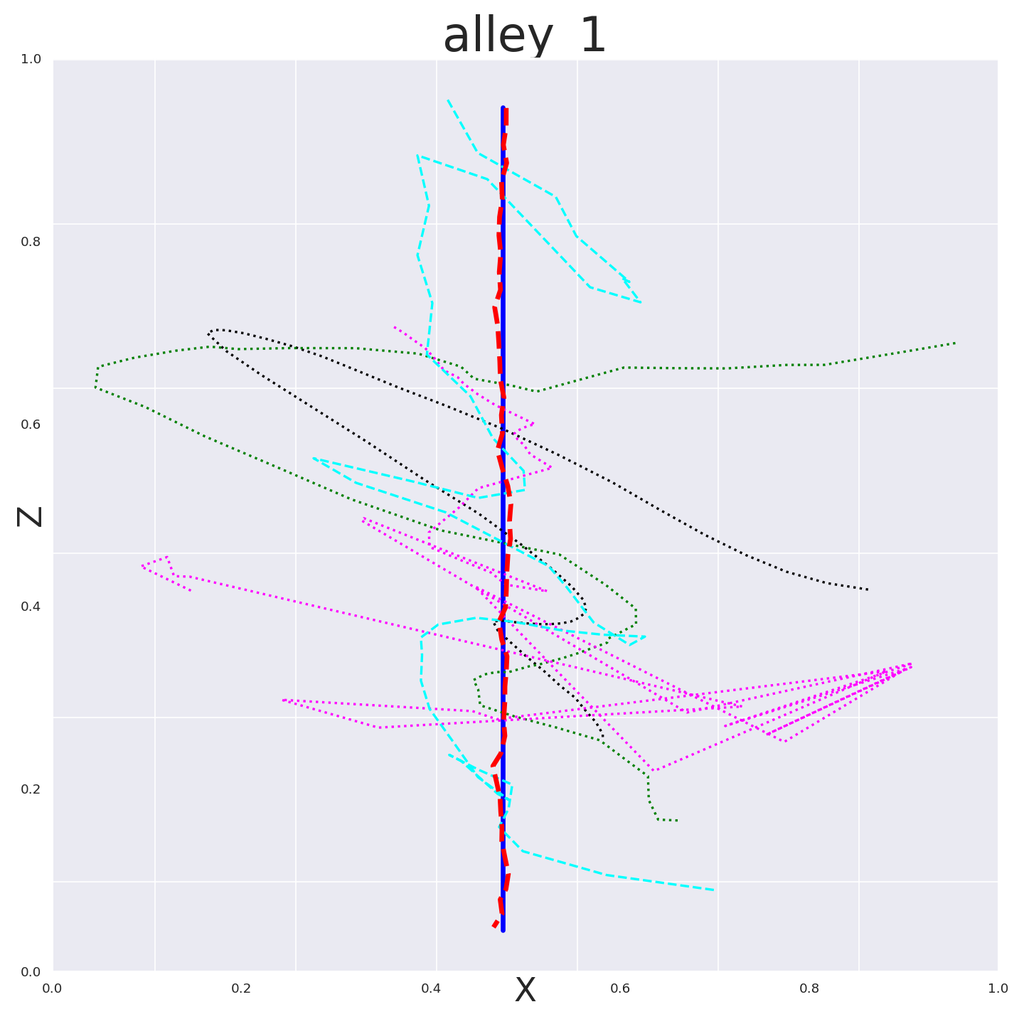} &
            \includegraphics[width=\itemwidth]{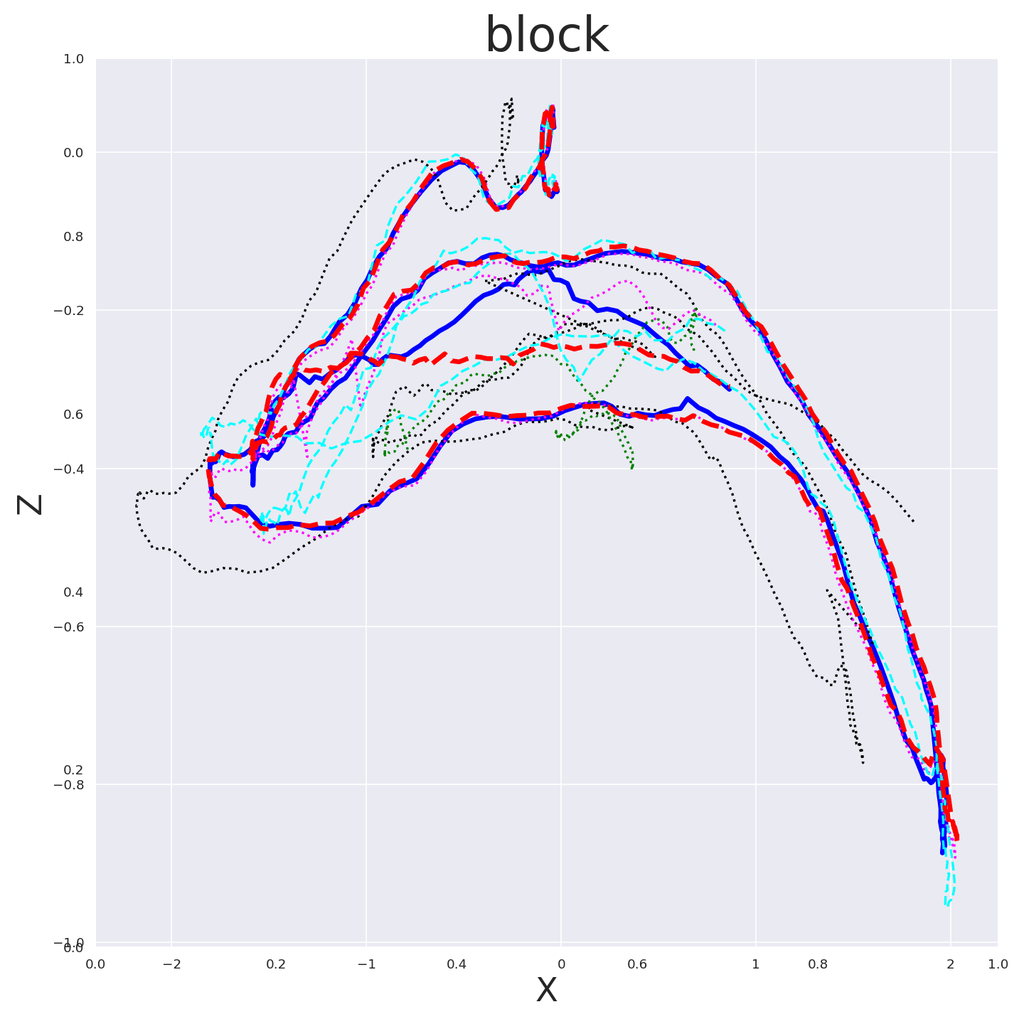} &
            \includegraphics[width=\itemwidth]{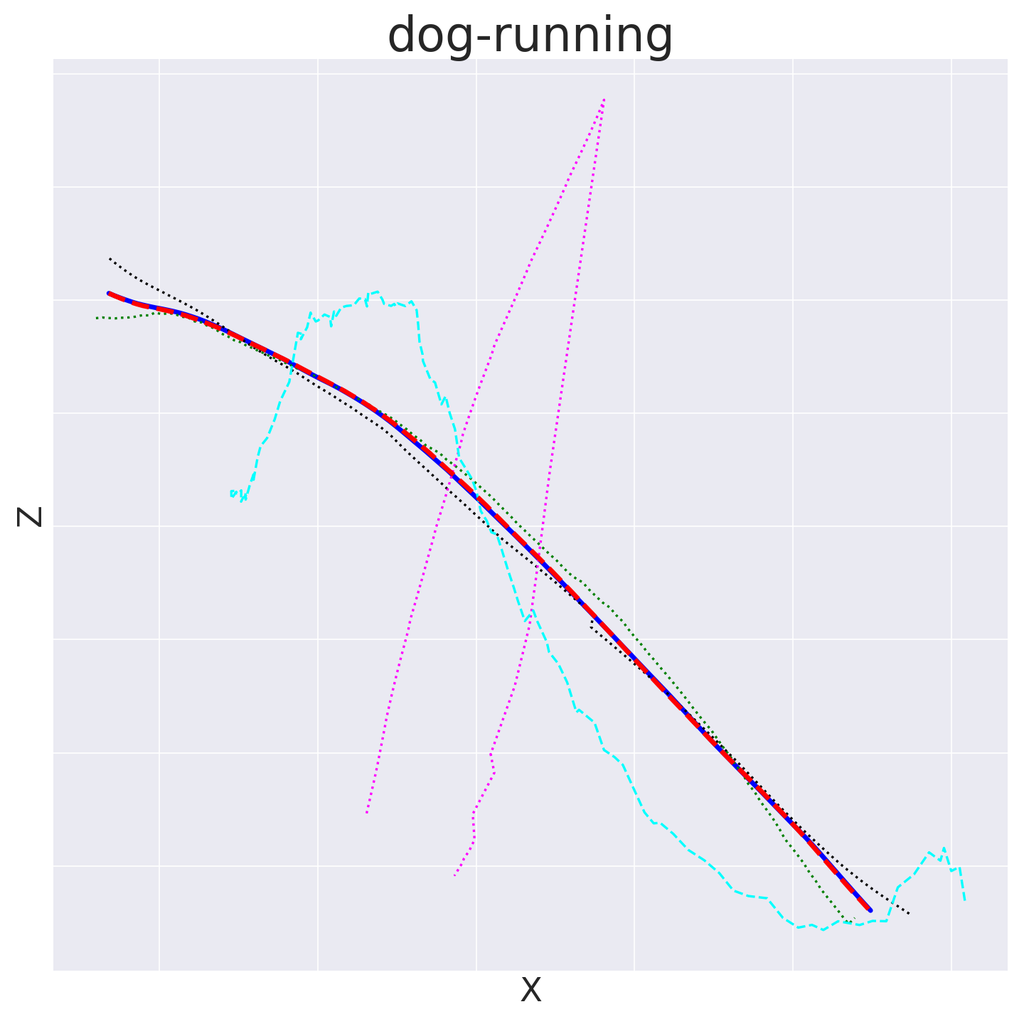} \\
            \includegraphics[width=\itemwidth]{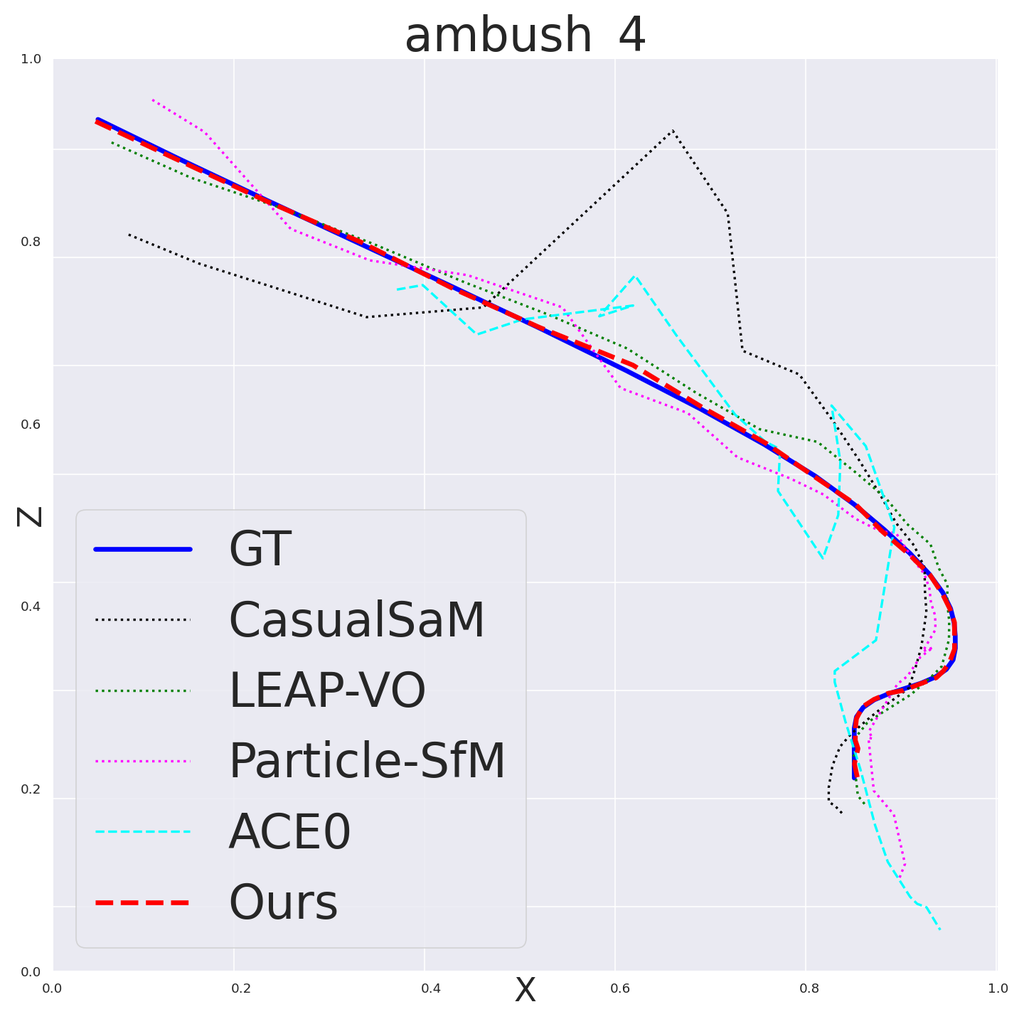} &
            \includegraphics[width=\itemwidth]{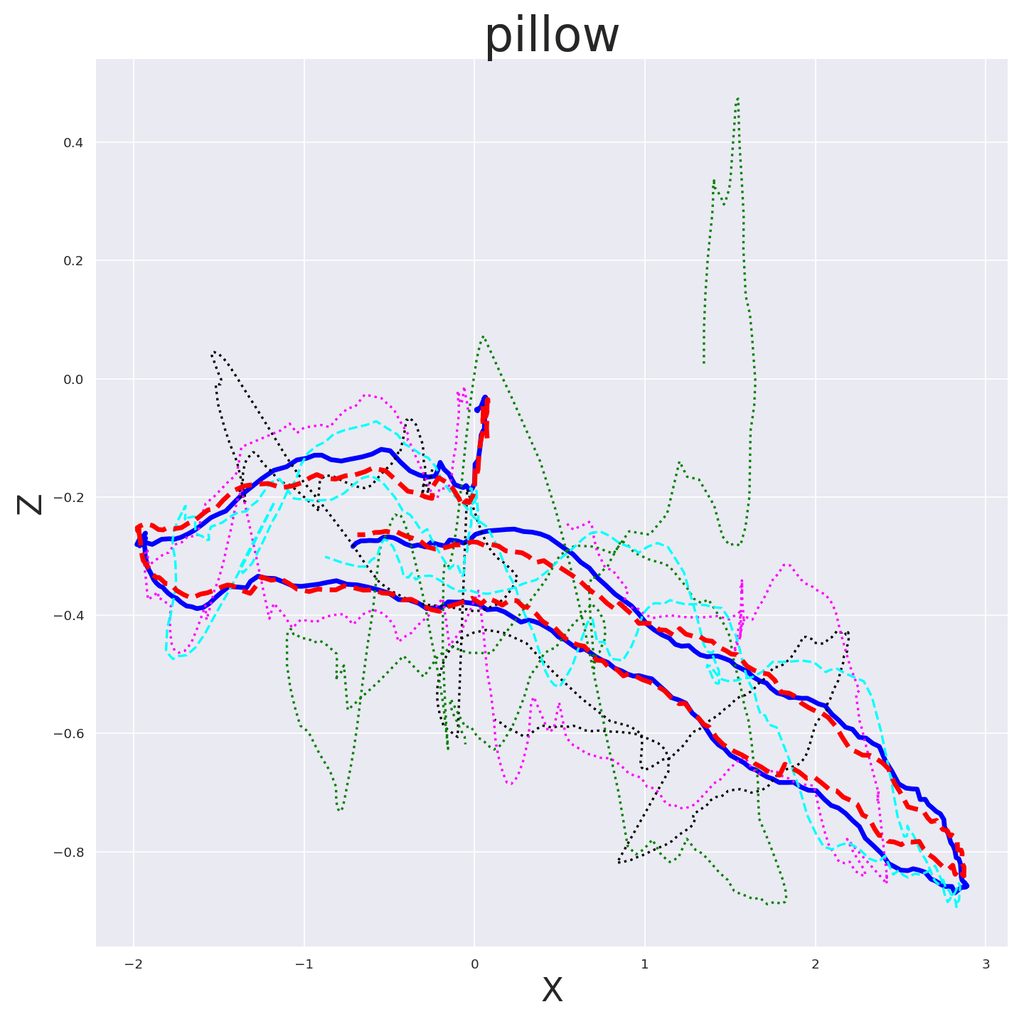} &
            \includegraphics[width=\itemwidth]{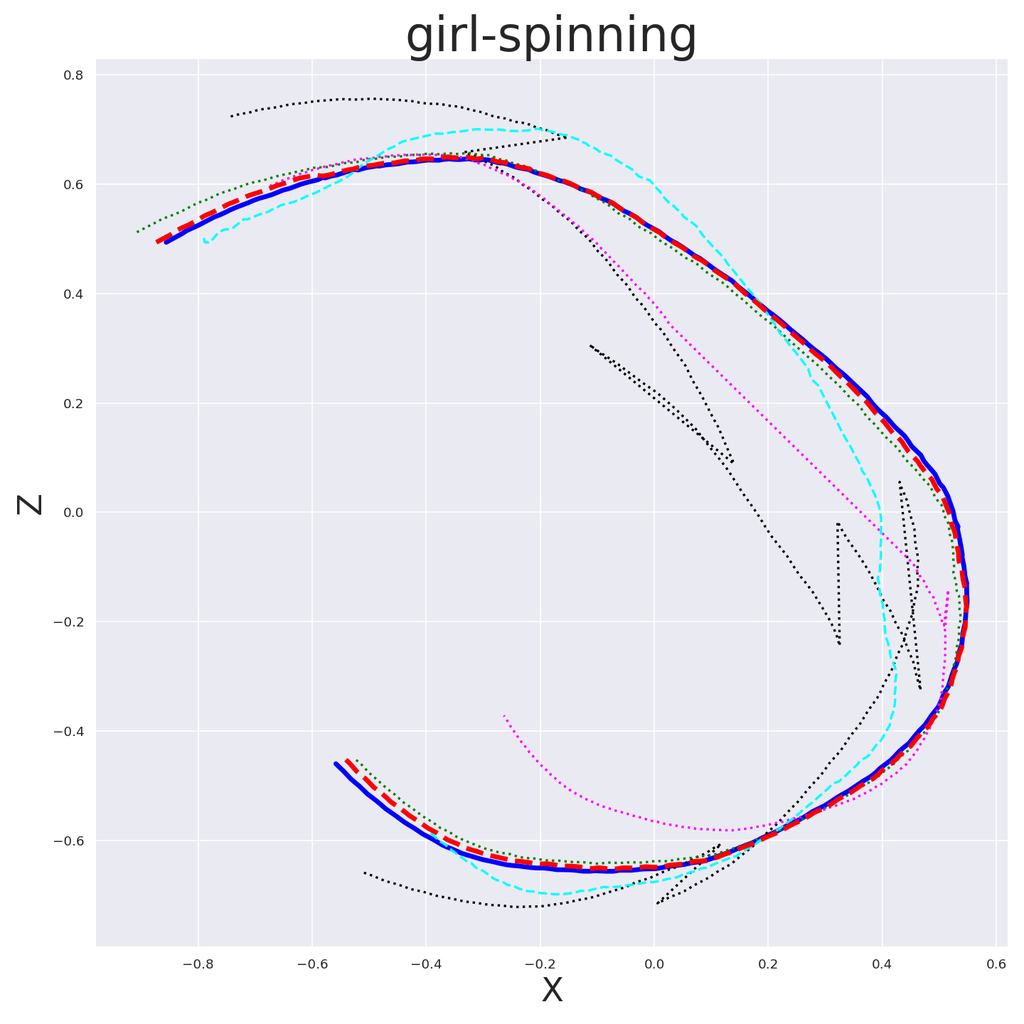} \\
            \footnotesize Sintel~\cite{Butler:ECCV:2012} & 
            \footnotesize Dycheck~\cite{gao2022monocular} &
            \footnotesize In-the-Wild~\cite{li2023dynibar} \\
    \end{tabular}
    \vspace{-0.5em}
     \caption{\textbf{Visualization of estimated camera trajectories.} Due to scene dynamics, our camera estimate (red dash) deviates less from the ground truth camera trajectory (blue solid line) than all other baselines. }
    \label{fig:cam-viz}
    \vspace{-0.5em}
\end{figure}

\medskip
\noindent \textbf{Metrics.}
We use standard error metrics to evaluate 
camera pose estimation: Absolute Translation Error (ATE), Relative Translation Error (RTE), and Relative Rotation Error (RRE). 
Following CasualSaM,
we normalize ground truth camera trajectories to be unit length, since camera tracks in different videos can vary significantly, and videos with longer trajectories can have higher influence on the computed metrics. 
For all methods, we align estimated camera paths to the ground truth trajectory by computing a global $\mathrm{Sim}(3)$ transformation via Umeyama alignment~\cite{umeyama1991least}. 
We report the average running time by dividing total running time of each method by the number of input frames. 
Additionally, we compare the quality of estimated video depths to recent baselines, adopting standard depth metrics: Absolute Relative
Error, log RMSE, and Delta accuracy. We
follow the standard evaluation protocol by excluding points that are further than 100 meters. For all methods, we align predicted video depths with ground truth through a global scale and shift estimate.

\subsection{Quantitative Comparisons}

Numerical results for camera pose estimation on the three benchmarks are reported in Tables~\ref{table:sintel}, \ref{table:dycheck}, and \ref{table:dynibar}. 
Our approach demonstrates significant improvement and achieves the best camera tracking accuracy on all error metrics in both the calibrated and uncalibrated setting, while being competitive in terms of running time. 
Notably, our approach outperforms the concurrent work MonST3R~\cite{zhang2024monst3r} in terms of both robustness and accuracy even though MonST3R adopts a more recent global 3D point cloud representation for dynamic scenes. 
In addition, we report results for depth prediction on the Sintel and Dycheck in Table~\ref{table:sintel_depth}. Our depth estimates again outperform other baselines significantly in all metrics.

\begin{figure}[tb]
\centering
  \includegraphics[width=1.0\columnwidth]{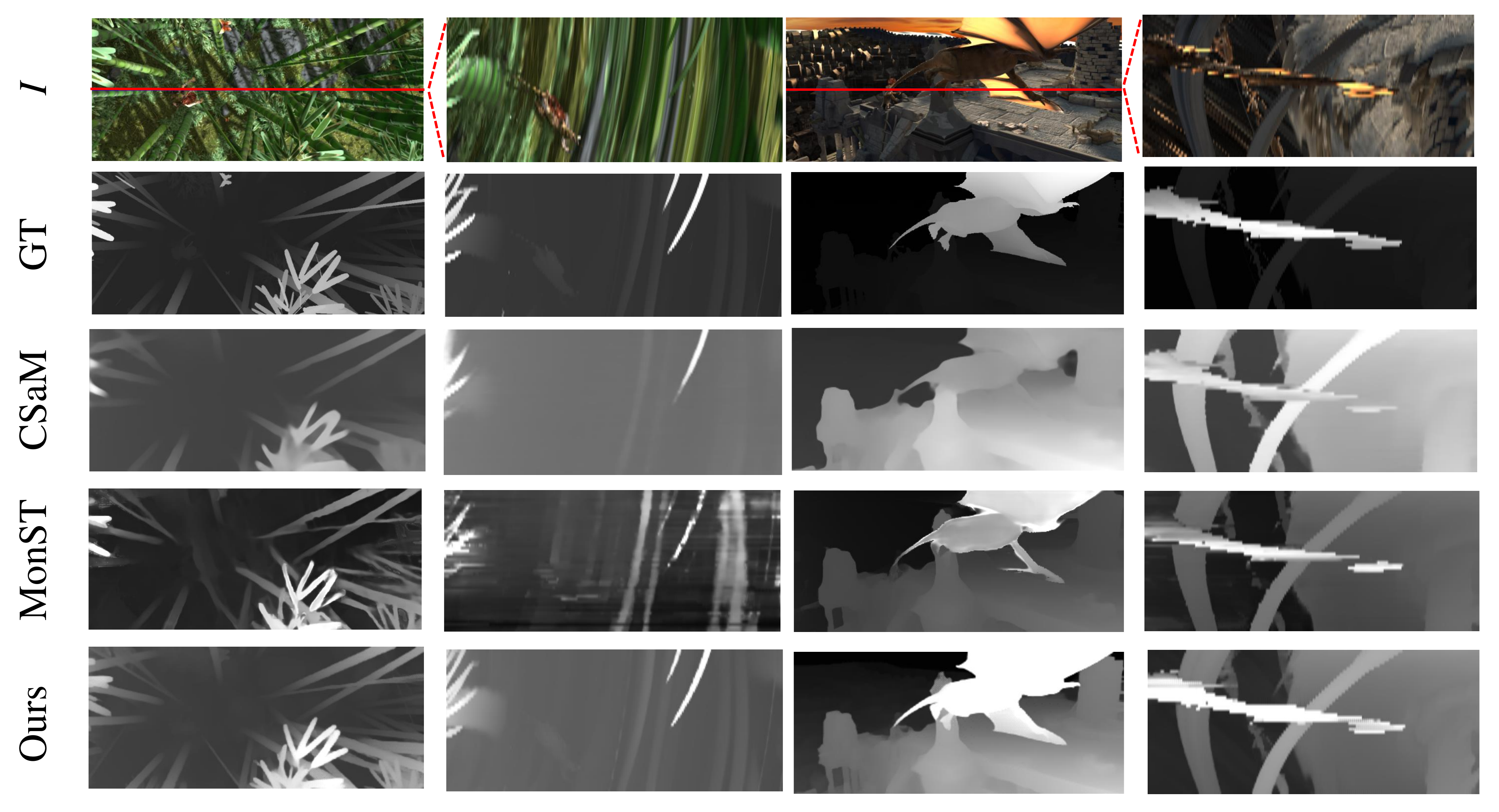} 
  \vspace{-1.em}
  \caption{\textbf{Visual comparisons of video depths.} We compare video depth estimates from our approach and from CasualSAM~\cite{zhang2022structure} and MonST3R~\cite{zhang2024monst3r} by visualizing their depth maps (odd columns) and corresponding $x$-$t$ slices (even columns). }
  \vspace{-1.em}
\label{fig:video_depth_viz}
\end{figure}

\setlength{\tabcolsep}{4pt}
\begin{table}[tb]
\begin{center}
\small
\begin{tabular}{l ccc @{\hskip 1em} cc}
\toprule
& \multicolumn{3}{c@{\hskip 1em}}{Poses} &
\multicolumn{2}{c@{\hskip 1em}}{Depth}
\\
Method & 
ATE & RTE & RRE & Abs-Rel & $\delta_{1.25}$
\\ 
\midrule
Droid-SLAM~\cite{teed2021droid} & 0.030 & 0.022 & 0.50 & - & -  \\
\midrule
w/o mono-init. & 0.038 & 0.026 & 0.49 & - & -  \\
w/o $\MotionMask$ & 0.032 & 0.127 & 0.14 & - & -\\
w/o 2-stage train. & 0.035 & 0.136 & 0.17 & - & - \\
w/o u-BA & 0.033 & 0.013 & 0.11 & - & - \\
\midrule
w/ ft-pose & 0.041 & 0.018 & 0.33 & 0.23 & 71.2 \\
w/o new $\mathcal{C}_{\text{prior}}$ & - & - & - & 0.36 & 72.5 \\
\bottomrule
Full & \textbf{0.019} & \textbf{0.008} & \textbf{0.04} & \textbf{0.21} & \textbf{73.1}
\end{tabular}
\caption{\textbf{Ablation study on the Sintel dataset.} Sec.~\ref{sec:ablation} describes each configuration.} \label{table:ablation} 
\end{center} 
\vspace{-1.5em}
\end{table}

\subsection{Ablation study} \label{sec:ablation}

We perform an ablation study to validate major design choices for our camera tracking and depth estimation modules. 
Specifically, we evaluate camera tracking results with different configurations: 1) vanilla Droid-SLAM~\cite{teed2021droid}, 2) without mono-depth initialization (w/o mono-init.), 3) without object movement map prediction (w/o $\MotionMask$), 4) directly train the model on dynamic videos without using proposed two-stage training scheme (w/o 2-stage train.), 5) always turn on mono-depth regularization during global BA (w/o u-BA). We also ablate two major design decisions for video depth estimation: 1) jointly refine camera poses and depth estimation (w/o ft-pose), and 2) use original mono-depth prior loss from CasualSAM instead of our proposed one (w/o new $\mathcal{C}_{prior}$).
As shown in Table~\ref{table:ablation}, our full system outperforms all other alternative configurations.

\begin{figure*}[tb]
\centering
  \includegraphics[width=2.0\columnwidth]{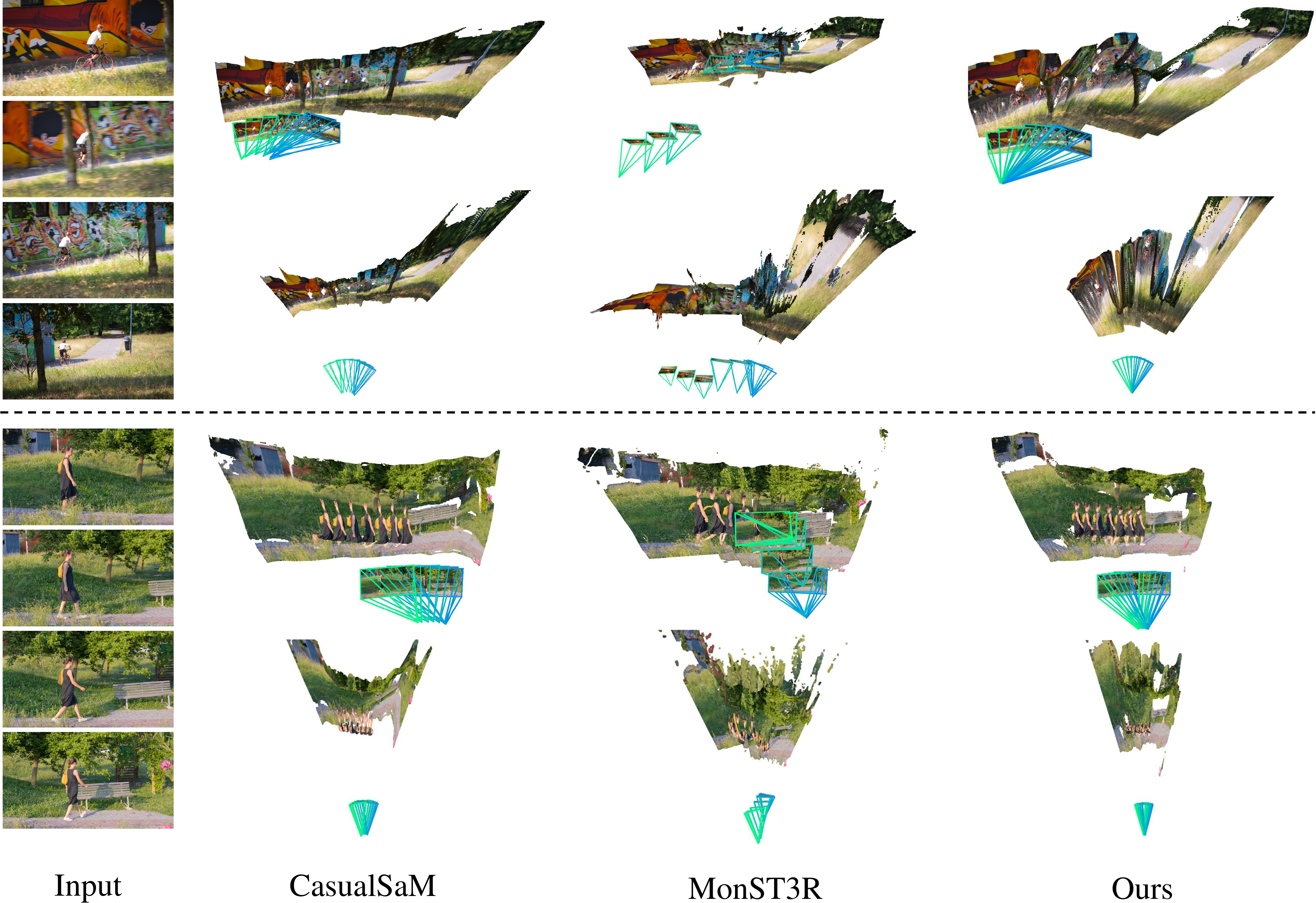} 
  \caption{We visualize two examples from DAVIS with rotation-dominant camera motion and narrow FoV. Compared to CasualSAM~\cite{zhang2022structure} and MonST3R~\cite{zhang2024monst3r}, our system produces more accurate camera and geometry estimates of underlying dynamic scenes.} %
\label{fig:Davis_vis}
\end{figure*}

\subsection{Qualitative Comparisons}
Fig.~\ref{fig:cam-viz} shows qualitative comparisons of estimated camera trajectory from our method and other baselines in 2D on three benchmarks, and our camera estimates  are the closest to the ground truth. Furthermore, we visualize and compare the estimated video depths from our approach and two recent optimization based techniques, CasualSAM~\cite{zhang2022structure} and MonST3R~\cite{zhang2024monst3r}, in Fig~\ref{fig:video_depth_viz}. In particular, we visualize both the depth map of a reference frame and corresponding $x$-$t$ depth slices over the entire video. Our approach once again produces more accurate, detailed and temporally consistent video depths.

Last, we compare reconstruction and camera tracking quality among different approaches on challenging examples from DAVIS~\cite{Perazzi2016} by visualizing their estimated cameras and unprojected depth maps. 
As shown in Fig.~\ref{fig:Davis_vis}, CasualSaM tends to produce distorted 3D point clouds, whereas MonST3R incorrectly treats rotational camera movement as translational one. 
In contrast, our approach produces more accurate camera along with more consistent geometry for such challenging inputs.

\section{Discussion and conclusion}

\paragraph{Limitations.} 
Despite excellent performance on a variety of in-the-wild videos, we observe that our approach can fail in extremely challenging scenarios, similar to findings from prior work~\cite{zhang2022structure}. For instance, camera tracking fails if moving objects dominate the entire image or if there is nothing for the system to track reliably. Please see supplemental material for visualization of failure cases.
Furthermore, our system cannot handle videos with varying focal lengths or strong radial distortion within the video.  Incorporating better priors from current vision foundation models into the pipelines is a promising future direction to explore.

\paragraph{Conclusion.}
We presented a pipeline that produces accurate camera parameters and consistent depths from casual monocular videos of dynamic scenes. Our approach efficiently scales to in-the-wild footage of varying time duration, with unconstrained camera paths and complex scene dynamics. We have shown that, with careful extension, prior deep visual SLAM and SfM frameworks can be extended to achieve strong generalization to a broad range of videos and outperform recent state-of-the-art methods significantly. %

{\small
\bibliographystyle{ieee_fullname}
\bibliography{refs}
}
\newpage
\appendix

\begin{figure*}[tb]
    \centering
    \setlength{\tabcolsep}{0.025cm}
    \setlength{\itemwidth}{4.2cm}
    \renewcommand{\arraystretch}{0.5}
    \hspace*{-\tabcolsep}\begin{tabular}{cccc}
            \includegraphics[width=\itemwidth]{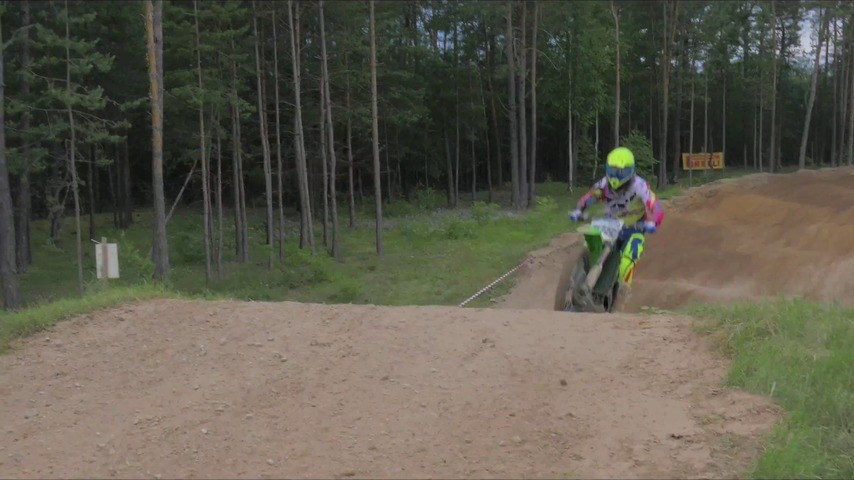}& 
            \includegraphics[width=\itemwidth]{figures/limitations/motor/00020.jpg}&
            \includegraphics[width=\itemwidth]{figures/limitations/motor/00039.jpg} &
            \includegraphics[width=\itemwidth]{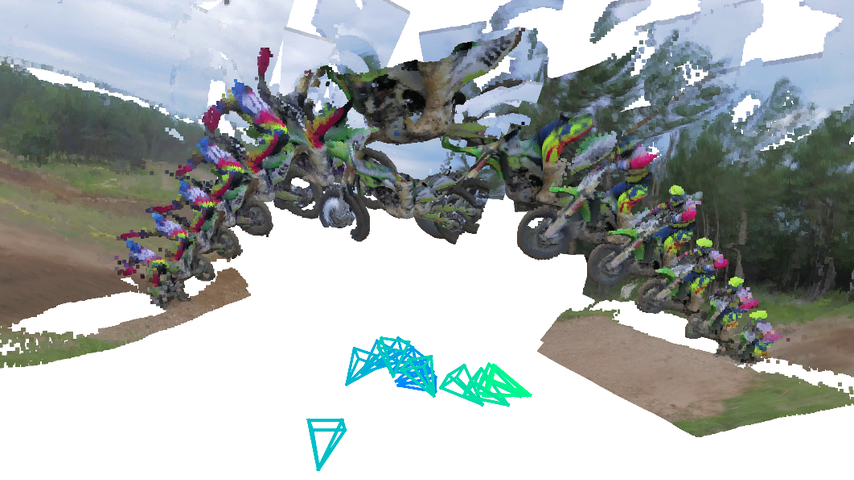} \\
            \includegraphics[width=\itemwidth]{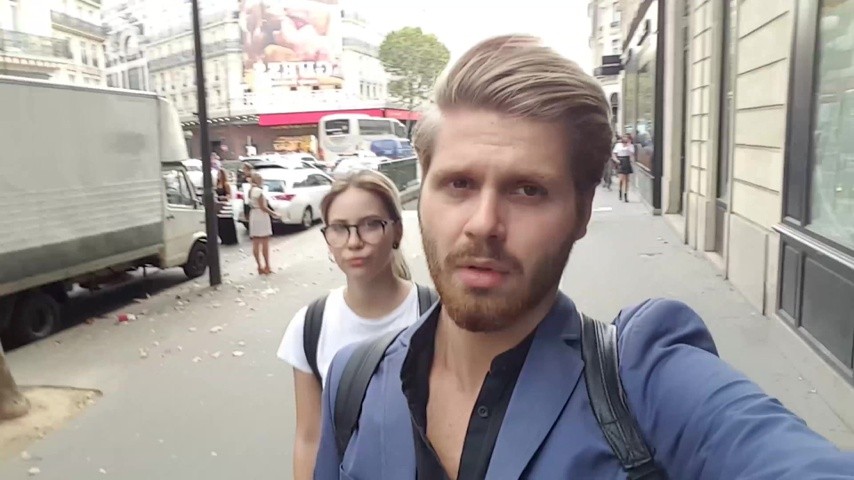} & 
            \includegraphics[width=\itemwidth]{figures/limitations/walking/00000.jpg} &
            \includegraphics[width=\itemwidth]{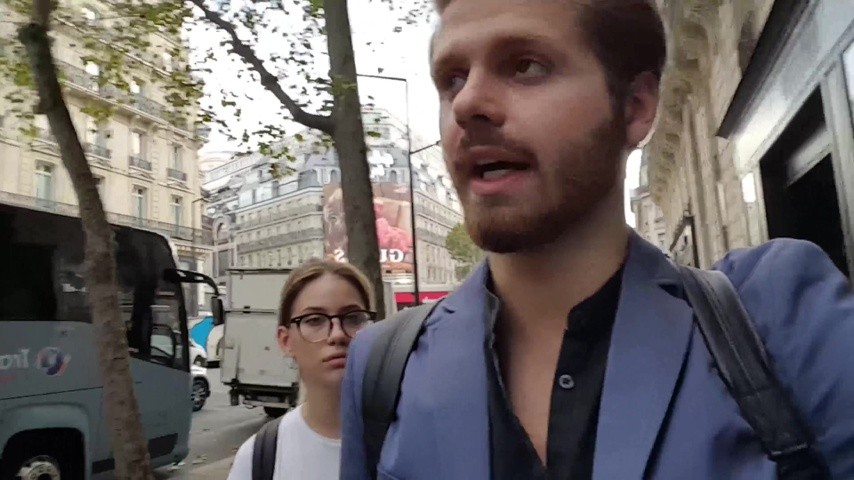} &
            \includegraphics[width=\itemwidth]{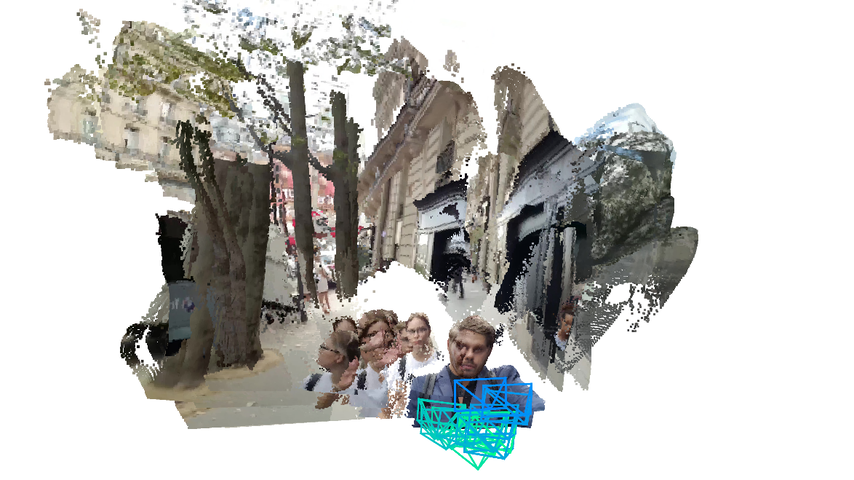}
    \end{tabular} 
    \vspace{-0.5em}
    \caption{\textbf{Limitations.} We visualize three reference video frames on the left and their corresponding estimated camera paths and reconstruction on the right. Our method can lose tracks in cases where a moving object dominates the entire videos (top row). Our approach can also struggle in cases where object motion and camera motion are colinear, such as the selfie video in the bottom row.}
    \label{fig:limitations}
\end{figure*}

\section{Implementation Details}

\begin{figure*}[tb]
\centering
  \includegraphics[width=2.0\columnwidth]{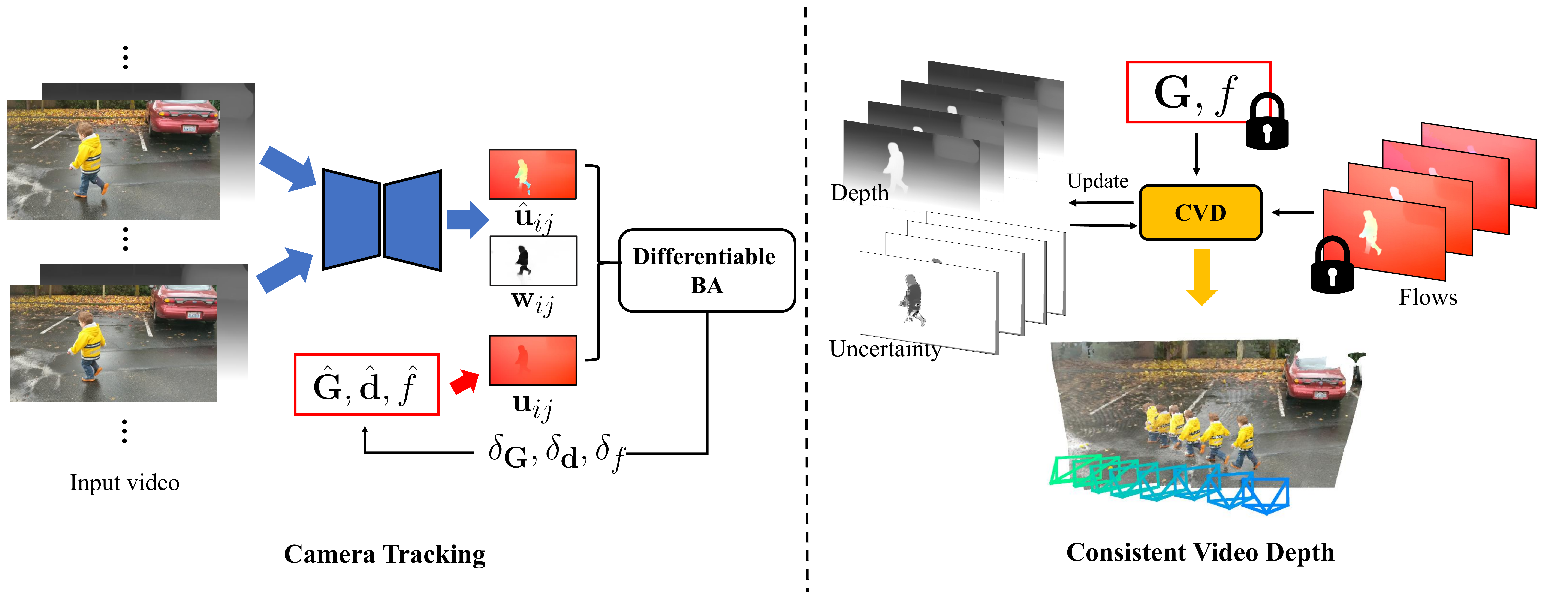} 
      \caption{\textbf{System overview.} \textbf{Left}:
      we estimate camera poses, focal length and low-resolution disparity maps from the input monocular video through differentiable Bundle Adjustment (BA): the network iteratively updates these state variables by learning to predict low-resolution flow $\hat{\mathbf{u}}_{ij}$, confidence, and movement probability maps $\mathbf{w}_{ij}$ and minimize weighted reprojection error between predicted flow ${\hat{\mathbf{u}}}_{ij}$ and flow induced by ego-motion ${\mathbf{u}}_{ij}$.
      We also initialize estimated disparity with mono-depth predicted from  off-the-shelf models~\cite{depthanything, piccinelli2024unidepth}. \textbf{Right}: we fix estimated camera parameters and perform first-order global optimization over video depth and corresponding uncertainty parameters by minimizing flow and depth losses through pairwise 2D optical flows.  
      } %
\label{fig:overview}
\end{figure*}

\subsection{System Overview}

Figure~\ref{fig:overview} shows an overview of our MegaSaM system. We separate the problem of camera and scene structure estimation into two stages, in the spirit of a conventional SfM pipeline~\cite{schonberger2016structure, schonberger2016pixelwise}. In particular, we first estimate camera poses $\Pose$, focal length $\focal$ and low-resolution disparity $\DroidDepth$ from the input monocular video through differentiable Bundle Adjustment (BA), where we initialize $\DroidDepth$ with monocular depth maps predicted from off-the-shelf models~\cite{depthanything, piccinelli2024unidepth}. In the second consistent video depth estimation phase, we fix estimated camera parameters and perform first-order optimization over video depth and uncertainty maps by enforcing flow and depth losses induced by pairwise 2D optical flows.

\subsection{Framework and Architecture}

We follow 
DROID-SLAM~\cite{teed2020raft} for feature extraction, correlation feature construction, and perform iterative BA updates through flow, confidence, motion probability predictions. Each input to the model is a pair of video frames $(I_i, I_j)$.

\begin{figure}[tb]
\centering
  \includegraphics[width=1.0\columnwidth]{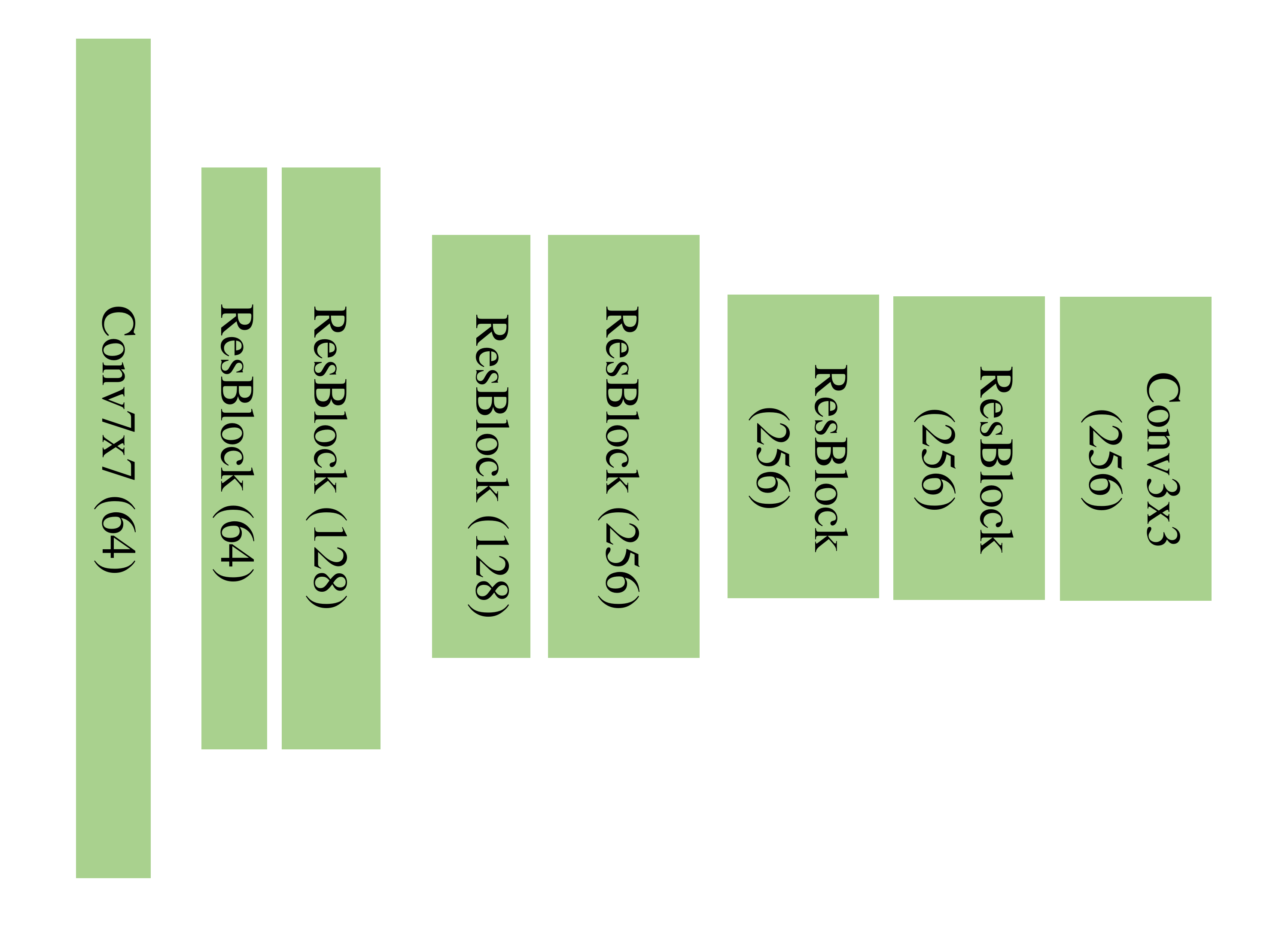} 
      \caption{Architecture of the feature and context encoders. Both encoders extract low-resolution features from input video frames at $\frac{1}{8}$ of the original resolution.} %
\label{fig:context}
\end{figure}

\medskip
\noindent \textbf{Feature extraction.}
We use context and feature encoders to encode each input video frame into two different low-resolution feature maps at $\frac{1}{8}$ resolution of the input image, as shown in Figure~\ref{fig:context}.

\medskip
\noindent \textbf{Correlation feature construction.}
The correlation layer constructs a 4D correlation volume from the features encoded from an image pair, and each  entry of the volume contains inner product of one pairs of feature vectors from the image pair.

\begin{figure*}[tb]
\centering
  \includegraphics[width=2.0\columnwidth]{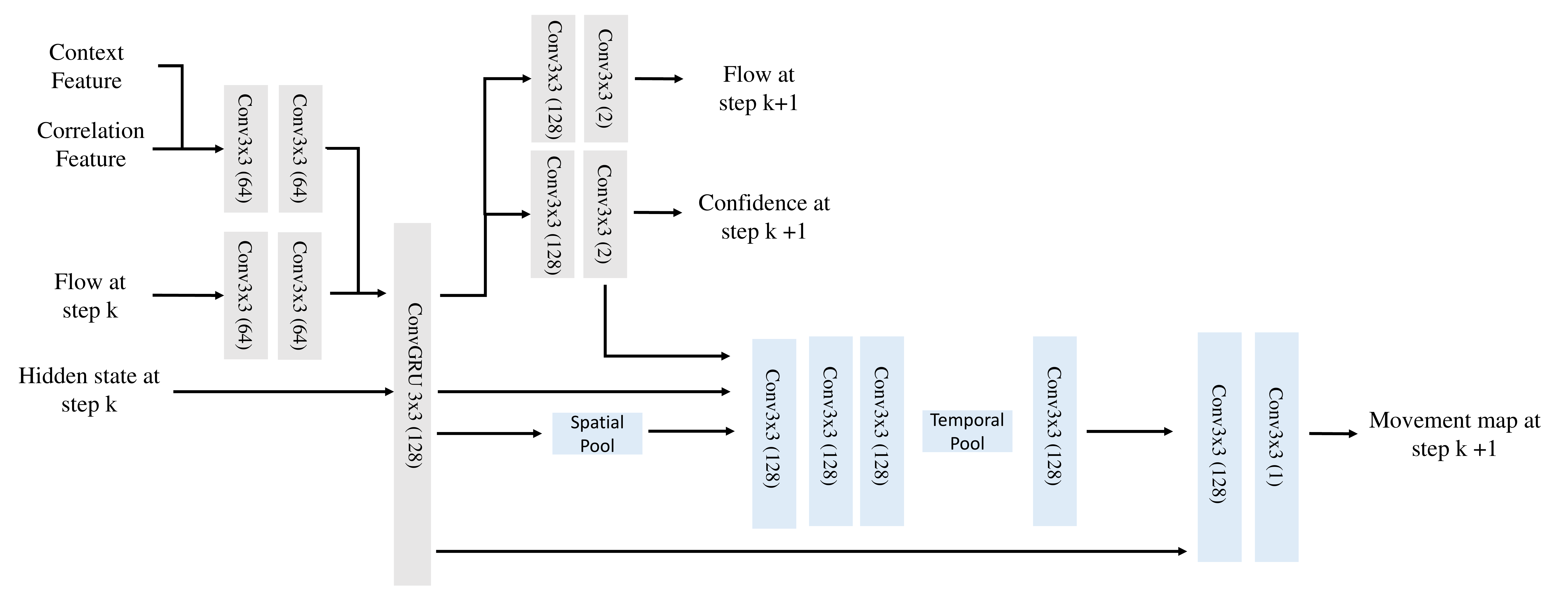} 
  \caption{Architecture of flow, confidence and movement map predictor. The gray blocks belong to the network $\RNN$ for flow and confidence prediction, and the blue blocks belong to the network $\RNN_m$ for object movement map prediction. In the first stage, we perform ego-motion pretraining for $\RNN$. In the second stage, we perform dynamic fine-tuning for $\RNN_m$ while fixing the parameters of $\RNN$.} 
\label{fig:gru}
\end{figure*}

\medskip
\noindent \textbf{Iterative updates.}
During each iterative BA step $k$, we update camera parameters and low-resolution disparity through flow, confidence and motion probability prediction. 
In particular, we first pretrain $\RNN$ on synthetic video data (ego-motion pretraining in the main paper) to learn to predict flows and corresponding flow confidence, as shown by the gray blocks in Figure~\ref{fig:gru}. In the second dynamic finetuning phase, we freeze the parameters of $\RNN$ and finetune the motion module $\RNN_m$ to predict extra object motion probability maps conditioned on the features from the ConvRGU, as shown in the blue blocks in Figure~\ref{fig:gru}. Within the motion module, we first perform 2D spatial average pooling to provide the model with global spatial information; we then perform average pooling along the time axis to fuse information from $I_i$ and all its neighboring keyframes $I_j$ (where $j \in \mathcal{N}(i)$).

\subsection{Consistent Video Depth Optimization}

Recall, from Section 3.3 of our main paper, that we follow 
CasualSAM~\cite{zhang2022structure} to estimate consistent video depth by performing an additional first-order optimization on video disparity $\hat{D}_i$ along with per-frame aleatoric uncertainty maps $\hat{M}_i$. Instead of jointly optimizing camera parameters and scene structure as in CasualSAM, however, we fix camera parameters as done in conventional SfM pipelines like COLMAP~\cite{schonberger2016structure, schonberger2016pixelwise}.

Our objective consists of three main cost functions:
\begin{equation}
    \mathcal{C}_{\text{cvd}} = w_{\text{flow}} \mathcal{C}_{\text{flow}} +  w_{\text{temp}} \mathcal{C}_{\text{temp}} + w_{\text{prior}} \mathcal{C}_{\text{prior}}
\end{equation}

We treat object motion in the video as the heteroscedastic aleatoric uncertainty of
the flow reprojection and depth consistency error~\cite{kendall2017uncertainties}, and assume the underlying noise is Laplacian~\cite{yang2020d3vo}. Specifically, for each selected pair $(I_i, I_j)$, flow reprojection loss $\mathcal{C}_{\text{flow}}$ compares $l_1$ loss weighted by the uncertainty $\hat{M}_{i}$ between flows $\text{flow}_{i \rightarrow j}$ from an off-the-shelf flow estimator~\cite{teed2020raft} and the correspondences $\FLowFields$ induced by our estimated camera motion and disparity through a multi-view constraint:
\begin{align}
    \mathcal{C}^{i \rightarrow j}_{\text{flow}} &= \hat{M}_{i} ||  \FLowFields - \mathbf{p}_i, \text{flow}_{i \rightarrow j}(\mathbf{p}_i) ||_1 + \log \left( \frac{1}{\hat{M}_{i}} \right), \\
    {\FLowFields} &= \pi \left( \Poseij \circ \pi^{-1} (\mathbf{p}_i, \hat{D}_i, K^{-1}) , K \right)
\end{align}
$\mathcal{C}_{\text{temp}}$ is an uncertainty weighted temporal depth loss that encourages pixel disparity to be temporally consistent according to estimated 2D optical flow:
\begin{align}
    \mathcal{C}^{i \rightarrow j}_{\text{temp}} &= \hat{M}_i \delta \left( \mathbf{P}^{i \rightarrow j}_z, \hat{D}_j(\mathbf{p} + \text{flow}_{i \rightarrow j}(\mathbf{p})) \right) + \log \left( \frac{1}{\hat{M}_{i}} \right) \nonumber \\
    \delta(a, b) &= || \max(\frac{a}{b}, \frac{b}{a}) ||_1 \nonumber \\
    \mathbf{P}^{i \rightarrow j}_z &= \left( D_i(\mathbf{p}) \mathbf{R}_{i \rightarrow j} \mathbf{K}^{-1} \mathbf{p} + \mathbf{t}_{i \rightarrow j} \right)_{[z]}
\end{align}
$\mathbf{R}_{i \rightarrow j}$ and $\mathbf{t}_{i \rightarrow j}$ are relative camera rotation and translation between $I_i$ and $I_j$; ${}_{[z]}$ is an operator that retrieve the third component of the 3D point vector (i.e. $z$ value).

$\mathcal{C}_{\text{prior}}$ is a depth prior loss that stops the final estimated video disparity from drifting too much from the initial estimate from the mono-depth network, and it consists of three losses:
\begin{align}
    \mathcal{C}_{\text{prior}} = \mathcal{C}_{\text{si}} + w_{\text{grad}} \mathcal{C}_{\text{grad}} + w_{\text{normal}} \mathcal{C}_{\text{normal}}
\end{align}

The scale-invariant depth loss $\mathcal{C}_{\text{si}}$ computes the mean square error of the difference among all pairs between optimized log-disparity $\log \hat{D}_i$ and initial log-disparity from the metric-aligned mono-depth prediction $\log D_i^{\text{align}}$.
\begin{align}
    \mathcal{C}_{\text{si}} &= \frac{1}{n} \sum_{(\mathbf{p})} (R(\mathbf{p}))^2 - \frac{1}{n^2} \left( \sum_{(\mathbf{p})} R(\mathbf{p}) \right)^2 \nonumber \\
    R_i &= \log (\hat{D}_i) - \log (D_i^{\text{align}}).
\end{align}

$\mathcal{C}_{\text{grad}}$ is a multi-scale scale-invariant gradient matching term~\cite{li2018megadepth}, which computes $l_1$ difference between estimated log disparity gradients and initial log-disparity gradients
\begin{align}
    \mathcal{C}_{\text{grad}} &= \frac{1}{n} \sum_s w_{\nabla}^{s}(\mathbf{p}) \sum_{\mathbf{p}} \left( |\nabla_x R^s (\mathbf{p})| + |\nabla_y R^s(\mathbf{p})| \right) \nonumber \\
    w_{\nabla}^{s}(\mathbf{p}) &= 1 - \exp \left( - \beta_{\nabla} (\nabla_x R^s(\mathbf{p}) + \nabla_y R^s(\mathbf{p})) \right)
\end{align}

\noindent where $R^s(\mathbf{p})$ is log-depth difference map at pixel position $\mathbf{p}$ and scale $s$. In other words, we only apply multi-scale gradient matching loss to pixels where the current estimated disparity deviates significantly from the original mono-depth. 

$\mathcal{C}_{\text{normal}}$ is a surface normal loss that encourages that normal $\hat{\mathbf{N}} (\mathbf{p})$ derived from estimated disparity to be close to the surface normal $\mathbf{N}^{\text{align}}$ derived from the initial metric-aligned monocular disparity:
\begin{align}
     \mathcal{C}_{\text{normal}} = \sum_{\mathbf{p}} 1 - \hat{\mathbf{N}} (\mathbf{p}) \cdot \mathbf{N}^{\text{align}} (\mathbf{p})
\end{align}

We set $w_{\text{grad}} = 1, w_{\text{normal}}=4, \beta_{\nabla}=5$ throughout our experiments. We simply choose image pairs $(I_i, I_j)$ from a set of fixed intervals following prior work~\cite{zhang2022structure}: $j \in (i + 1, i + 2, i + 4, i + 8, i + 15)$. During optimization, we initialize the disparity variables from the metric-aligned monocular depth by combining estimates from off-the-shelf modules as described in the main paper~\cite{depthanything, piccinelli2024unidepth}, and we initialize the uncertainty map with object motion probability maps predicted from our camera tracking module.
The optimization first conducts a ``warm-up'' phase for 100 steps by fixing the video disparity variables and optimizing the per-frame uncertainty map, per-frame scale, shift variables using the aforementioned losses. The disparity maps and uncertainty maps are then optimized together under the aforementioned losses for another 400 steps.

\subsection{Additional Details}

\noindent \textbf{Training Losses.}
We supervise our network using a combination of pose loss and flow loss. The flow
loss is applied to pairs of adjacent frames. We compute the optical flow induced by the predicted
depth and poses and the flow induced by the ground truth depth and poses. The loss is taken to be the
average l2 distance between the two flow fields.

Given a set of ground truth poses $\{ \mathbf{T}_i \}_{i=1}^N$ and predicted poses $\{ \mathbf{G}_i \}_{i=1}^N$, the pose loss is taken to be the distance between the ground truth and predicted poses,  $\mathcal{L}_{pose} = \sum_i || \text{Log}_{SE(3)}(\mathbf{T}_i^{-1} \cdot \mathbf{G}_i) ||_2$. We apply the losses to the output of every BA iteration with exponentially increasing weight using $\gamma = 0.9^{k}$, where $k$ indicates the $k^{th}$ BA iterations.

\medskip
\noindent \textbf{Training and Inference Details}
In our two-stage training scheme, we first pretrain our model on synthetic data of static scenes,
which include 163 scenes from TartanAir~\cite{tartanair2020iros} and 5K videos from static Kubric~\cite{greff2021kubric}. In the second stage, we finetune motion module $\RNNm$ on 11K dynamic videos from Kubric~\cite{greff2021kubric}. 
Each training example consists of a 7-frame video sequence. We first precompute a distance matrix between each pair of video frame based on the average ego-motion induced flow magnitude. We then dynamically generate a training sequence according to the constructed distance matrix, we randomly sample each frame such that average flow between them is between 0.5px and 64px.

Within the camera tracking module, we normalize video disparity $\DroidDepth$ such that its $98$ percentile is $2$; we also normalize focal length by dividing it by the input image resolution within every the bundle adjustment stage.

\section{Limitations}

Despite excellent performance on a variety of in-the-wild videos, we observe that our approach can fail in extremely challenging scenarios, similar to findings from prior work~\cite{zhang2022structure}. For instance, camera tracking fails if moving objects dominate the entire image or if there is nothing for the system to track reliably, as shown in the first row of Fig.~\ref{fig:limitations}. Furthermore, our approach also struggles on dynamic videos where camera motion and object motion are colinear, as shown in the second row of Fig.~\ref{fig:limitations}.

\end{document}